%% file: main.tex
\documentclass[letterpaper]{article}

\usepackage{aaai2026}  

\usepackage{times}  
\usepackage{helvet}  
\usepackage{courier}  
\usepackage[hyphens]{url}  
\usepackage{graphicx} 
\usepackage{natbib}  
\usepackage{caption} 
\usepackage{amsmath}
\usepackage{pifont}   
\usepackage{amssymb}
\usepackage[table]{xcolor} 
\usepackage{relsize}
\usepackage{multirow}
\usepackage{adjustbox}
\usepackage{tabularx}
\usepackage{subcaption} 
\usepackage[hidelinks]{hyperref}

\usepackage{booktabs}    
\usepackage{multirow}    
\usepackage{arydshln}    
\usepackage{makecell}    
\usepackage{rotating}    
\usepackage{xcolor}      

\usepackage[ruled,vlined]{algorithm2e} 

\usepackage{algorithm}

\pdfoutput=1
\usepackage{arydshln}
\definecolor{lightblue}{RGB}{173,216,230}  
\definecolor{lightgreen}{RGB}{240, 255, 240}
\definecolor{verylightgray}{RGB}{240, 240, 240}
\usepackage{xcolor}
\definecolor{darkgreen}{rgb}{0.0, 0.4, 0.0}
\definecolor{darkred}{rgb}{0.75, 0.0, 0.0}
\definecolor{green_recursive}{HTML}{D7F5E1}
\definecolor{purple_recursive}{HTML}{F6E2F4}
\definecolor{blue_recursive}{HTML}{DCEAF7}
\definecolor{yellow_recursive}{HTML}{FFFFD5}
\newcommand{\ccol}{\cellcolor{blue!7}}

\title{Adapt-As-You-Walk Through the Clouds: Training-Free Online Test-Time
Adaptation of 3D Vision-Language Foundation Models}
\author{
    Mehran Tamjidi \equalcontrib \textsuperscript{\rm 1 \rm},
    Hamidreza Dastmalchi \equalcontrib \textsuperscript{\rm 2},
    Mohammadreza Alimoradijazi \textsuperscript{\rm 3},\\
    Ali Cheraghian\textsuperscript{\rm 4},
    Aijun An\textsuperscript{\rm 2},
    Morteza Saberi\textsuperscript{\rm 1}
}

\affiliations{
   \textsuperscript{\rm 1}School of Computer Science, University of Technology Sydney, Sydney, Australia\\
    \textsuperscript{\rm 2}Department of Electrical Engineering and Computer Science, York University, Toronto, Canada\\

     \textsuperscript{\rm 3}Business school, The University of New South Wales, Sydney, Australia\\
      \textsuperscript{\rm 4}School of Engineering, Macquarie University, Sydney, Australia\\

    [mehran.tamjidi, morteza.saberi]@uts.edu.au, [hrd, aan]@yorku.ca, reza.moradi@unsw.edu.au, ali.cheraghian@mq.edu.au
    
}

\usepackage{bibentry}

\begin{document}

\maketitle

\input{sec/0_abstract}


\input{sec/1_intro}

\input{sec/2_related_work}

\input{sec/3_method}

\input{sec/4_experiments}

\input{sec/5_Ablation_Study}

\input{sec/6_conclusion}

\bibliography{aaai2026}

\input{sec/X_suppl.tex}

\end{document}

%% file: sec/0_abstract.tex
\begin{abstract}
3D Vision-Language Foundation Models (VLFMs) have shown strong generalization and zero-shot recognition capabilities in open-world point cloud processing tasks. However, these models often underperform in practical scenarios where data are noisy, incomplete, or drawn from a different distribution than the training data.
To address this, we propose \textbf{Uni-Adapter}, a novel training-free online test-time adaptation (TTA) strategy for 3D VLFMs based on dynamic prototype learning. We define a 3D cache to store class-specific cluster centers as prototypes, which are continuously updated to capture intra-class variability in heterogeneous data distributions. These dynamic prototypes serve as anchors for cache-based logit computation via similarity scoring. Simultaneously, a graph-based label smoothing module captures inter-prototype similarities to enforce label consistency among similar prototypes. Finally, we unify predictions from the original 3D VLFM and the refined 3D cache using entropy-weighted aggregation for reliable adaptation. Without retraining, Uni-Adapter effectively mitigates distribution shifts, achieving state-of-the-art performance on diverse 3D benchmarks over different 3D VLFMs—improving ModelNet-40C by \textbf{10.55\%}, ScanObjectNN-C by \textbf{8.26\%}, and ShapeNet-C by \textbf{4.49\%} over the source 3D VLFMs. \textbf{Project page: \href{https://mehran-tam.github.io/Uni-Adapter}{https://mehran-tam.github.io/Uni-Adapter}}.
\end{abstract}

%% file: sec/1_intro.tex
\section{Introduction}
\label{sec:intro}

3D Vision-Language Foundation Models (VLFMs), such as Uni3D \cite{zhou2023uni3d}, have introduced remarkable potential in multimodal point cloud processing tasks. Pretrained on web-scale text-image-point cloud triplets, these models learn cross-modal representations in a shared embedding space, enabling zero-shot recognition of novel point cloud categories. Despite the strength of these models, VLFMs encounter critical limitations in real-world scenarios where acquired point clouds often suffer from severe noise, sparsity, and low resolution due to sensor constraints and environmental factors. Domain adaptation and generalization aim to address these distribution shifts by bridging the gap between source and target domains.

\begin{figure}[t!]
	\centering
	\captionsetup{skip=6pt} 
	\includegraphics[width=1
    \linewidth]{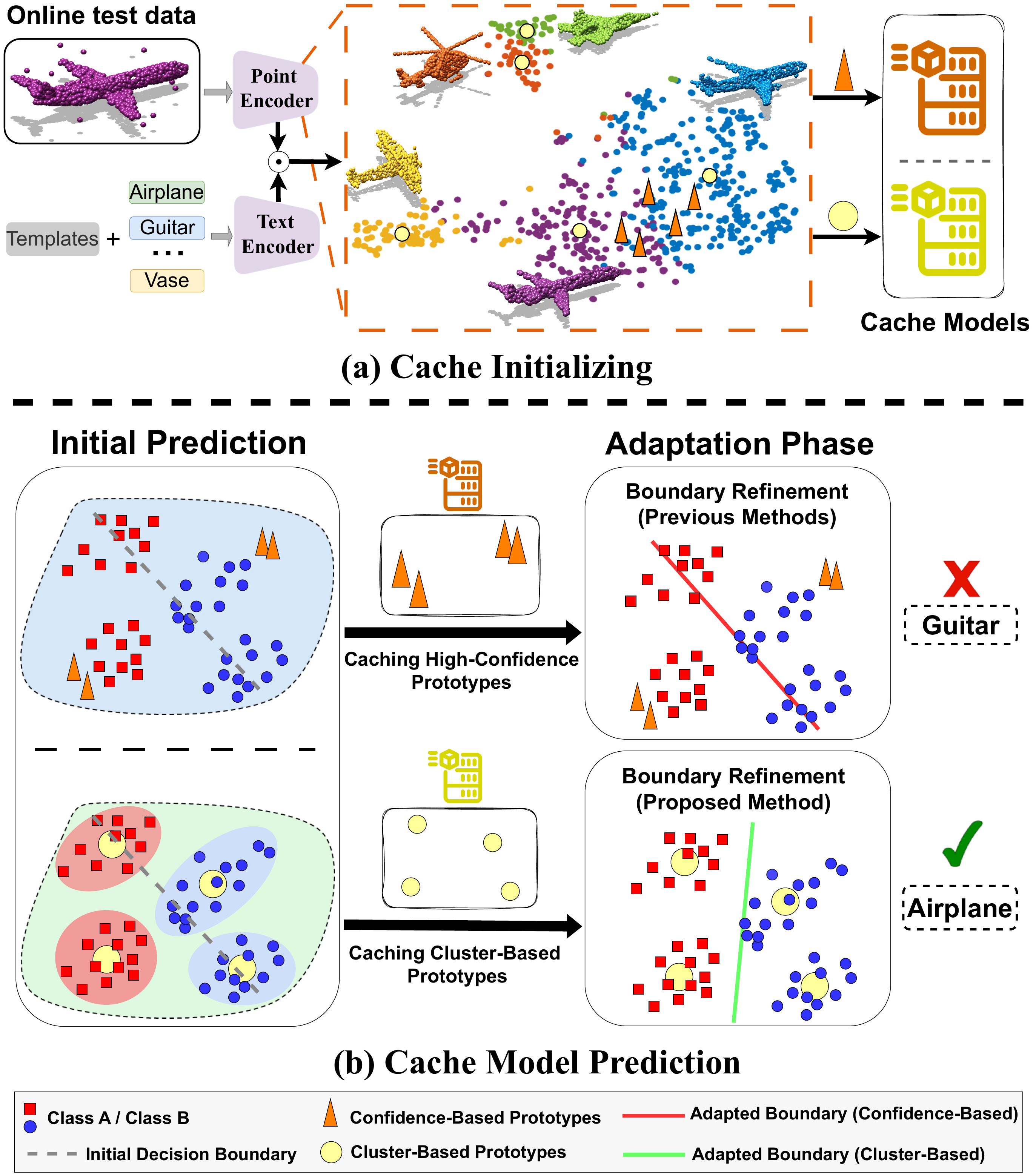}
\caption{(a) t-SNE of Uni3D embeddings for the airplane class in ModelNet40-C shows clear intra-class clustering patterns. Confidence-based prototypes (triangles) cache only high-confidence samples, while cluster-based prototypes (circles) represent distribution modes via online clustering.
(b) In the toy example, confidence-based caching leads to incorrect boundaries due to poor mode coverage, whereas cluster-based caching captures diverse patterns and enables correct predictions.}

    \vspace{-0.4cm}
	\label{fig:intro}
\end{figure}

Among existing adaptation approaches for VLFMs, test-time adaptation (TTA) ~\cite{o-tpt, TDA, huang2025cosmic} offers a particularly efficient solution, requiring no labeled target data while enabling dynamic adjustment to unseen conditions. Existing TTA methods for VLFMs can be broadly categorized into training-based and training-free approaches. Training-based TTA methods adapt to the target domain by updating a subset of model parameters \cite{2024watt} or soft prompts \cite{tpt, C-TPT, o-tpt} at test time. These methods typically optimize objectives like prediction entropy minimization across augmented views of test samples ~\cite{tpt}, or employ additional auxiliary losses~\cite{o-tpt} to guide adaptation. While effective in reducing domain shift, these methods often require iterative backpropagation, making them computationally demanding and less suited for real-time deployment. In contrast, training-free TTA methods, such as recent cache-based approaches~\cite{TDA, Point-Cache}, avoid parameter tuning by dynamically caching high-confidence features. These embeddings refine predictions via feature similarity, enabling lightweight, scalable adaptation for real-time and streaming scenarios.

While mainly designed for 2D VLFMs \cite{clip}, cache-based strategies are underexplored in 3D VLFMs, with only a few early attempts to design effective cache modules. Recently, Point-Cache~\cite{Point-Cache} introduced a TTA framework for 3D VLFMs, proposing a dual-cache structure of global and local caches. Both caches are built from high-confidence test samples, assuming these samples sufficiently represent the full data distribution. However, this assumption is often violated in practice—particularly for 3D data—where each semantic class can exhibit significant structural diversity. As illustrated in Figure~\ref{fig:intro}(a), features corresponding to a single class (e.g., ``airplane") form multiple distinct clusters in the feature space, reflecting different structural modes. Consequently, high-confidence prototypes typically capture only a subset of these variations, leading to suboptimal adaptation performance. This limitation is shown in the top of Figure~\ref{fig:intro}(b), where cached high-confidence prototypes (triangular markers) lead to incorrect decision boundaries.

To overcome the limitations of prior confidence-based cache strategies, we propose Uni-Adapter (Unified 3D Adapter), a novel online TTA framework for 3D VLFMs. Our approach employs a cluster-based caching strategy that dynamically stores and updates cluster centers, ensuring comprehensive coverage of the underlying feature distribution. This design is visualized in the lower part of Figure~\ref{fig:intro}(b), where yellow circular markers denote the cached cluster centers used as prototypes. These prototypes offer a more faithful representation of the distribution, leading to improved affinity calculations and more robust adaptation.

To enable clustering-based prototyping in the test-time setting, we adopt an online clustering strategy, where incoming test samples incrementally update class-specific cluster centers, serving as prototypes. Each class maintains multiple cluster-based prototypes within a unified cache, ensuring comprehensive coverage of diverse data distribution modes. This strategy captures intra-class variability and prevents over-reliance on a few dominant patterns. Moreover, we observe that the performance of existing cache-based models is affected by noisy pseudo-labels, allowing misclassified samples to contaminate the cache. To address this issue, we construct a similarity graph over cached prototypes and apply graph-based label smoothing to refine their labels.
This enables effective label propagation across similar prototypes, mitigating noisy pseudo-labels and yielding a more reliable, adaptive cache. We solve the resulting Laplacian system using the conjugate gradient method for its efficiency and scalability for large, sparse systems. Finally, we fuse the original 3D VLFM scores and 3D cache logits using entropy-driven confidence weighting to derive the final prediction.

In summary, the contributions of the proposed method are as follows: \textbf{1)} We introduce a cluster-based caching strategy that employs multiple cluster centers per class to capture intra-class variability, enabling adaptation to diverse test distributions. \textbf{2}) We apply graph-based label smoothing over cache prototypes, using inter-prototype similarities to refine noisy pseudo-labels and improve cache-based adaptation under distribution shift.  \textbf{3}) We conduct extensive experiments to validate our approach on different 3D VLFMs across diverse benchmarks—including corrupted datasets such as ShapeNet-C~\cite{mirza2022mate}, ModelNet-40C~\cite{modelnet40c}, and ScanObjectNN-C~\cite{mirza2022mate}, as well as clean datasets (large-scale and small-scale)—achieving new state-of-the-art results.

%% file: sec/2_related_work.tex
\vspace{-0.2cm}

\section{Related Work}

\noindent\textbf{3D Vision-Language Foundation Models (3D VLFMs)} have demonstrated transformative potential in advancing point cloud understanding by bridging semantic representations from large-scale image-text datasets to 3D data~\cite{pointclipv2, chen2023clip2scene, ulip2}. For instance,  Uni3D ~\cite{zhou2023uni3d}, ULIP ~\cite{ulip}, ULIP-2 ~\cite{ulip2}, and OpenShape~\cite{openshape} employ contrastive learning on extensive datasets of paired image, text, and point cloud data to achieve robust cross-modal feature alignment. These pre-trained 3D VLFMs exhibit strong zero-shot capabilities and geometric semantic perception across diverse tasks. However, their performance is often hindered by domain gaps, limiting generalization to real-world and dynamic scenarios.

\vspace{0.1cm}

\vspace{0.1cm}
\noindent\textbf{Test-Time Adaptation (TTA)}  focuses on dynamically adapting model predictions to novel domains without requiring target annotations or access to the source data \cite{sar, lame}.
Early TTA methods, designed for vision-only models, adapted parameters via post-hoc regularization during inference. For instance, TENT \cite{Tent}, SHOT \cite{shot}, and MEMO \cite{memo} minimize the entropy of the softmax prediction distribution to boost confidence and  generalization to the downstream domains. With advancements in VLFMs, recent TTA approaches leverage text modalities to enhance generalization. TPT \cite{tpt} and DiffTPT \cite{difftpt} combine entropy minimization with fine-tuning a learnable prompt for each test sample. SCAP \cite{zhang2025scap} optimizes both image and text prompts for TTA. While effective, these methods require costly gradient backpropagation at the test time. In contrast, TDA~\cite{TDA}, COSMIC~\cite{huang2025cosmic}, and PointCache~\cite{Point-Cache} use cached high-confidence prototypes to refine VLFM predictions through similarity-based scoring. 
However, relying solely on confident samples can miss distribution modes, and noisy prototypes can lead to suboptimal performance.

\vspace{0.1cm}

\noindent\textbf{Test-Time Point Cloud Adaptation} has gained significant traction in improving the generalization of 3D point cloud analysis across tasks, including recognition~\cite{Point-Cache, wang2024backpropagation, shim2024cloudfixer}, segmentation~\cite{zhao2025d, zou2024hgl}, registration~\cite{hatem2023point}, object detection~\cite{lin2024monotta, chen2024mos, yuan2024reg}, and scene completion~\cite{jang2025talos}. These approaches can be divided into two distinct groups. The first group modifies model parameters and employs training during inference. For instance, MATE~\cite{mirza2022mate} adapts encoder parameters through self-training, and Bahri et al.~\cite{bahri2024test} adapt normalization layers using TENT~\cite{wang2020tent}. The second group employs methods that avoid parameter updates. Specifically, BFTT3D~\cite{wang2024backpropagation} integrates source representations using a non-parametric adapter, whereas CloudFixer~\cite{shim2024cloudfixer} and 3DD-TTA~\cite{dastmalchi2024test} adapt input point clouds through geometric transformations guided by diffusion models. However, these methods are often designed for smaller-scale models and face challenges when applied to large, multi-modal 3D models. PointCache~\cite{Point-Cache}, closely related to our work, adapts VLFMs using global and local caches built from high-confidence predictions and applies k-means to summarize local patch features. In contrast, our Uni-Adapter performs online, confidence-weighted clustering at the class level to capture diverse distribution modes in 3D data.

%% file: sec/3_method.tex
\vspace{-0.1cm}
\section{Proposed Method}
\vspace{-0.2cm}
\subsection{Background}

3D VLFMs~\cite{ulip2, openshape, zhou2023uni3d} use separate encoders to map point clouds, images, and text into a shared, aligned feature space. A text encoder \(\mathrm{E}_\mathrm{T}\), typically based on CLIP~\cite{radford2021learning}, encodes class prompts, while a transformer-based point encoder \(\mathrm{E}_\mathrm{P}\), adapted with a point tokenizer, encodes 3D point clouds. In zero-shot classification, a generic prompt \(r = \text{"a point cloud of a"}\) is prepended to the \(i\)th class name \(y_i \in \mathcal{Y}\), where \(\mathcal{Y} = \{y_1, \dots, y_K\}\) denotes the set of \(K\) class names. The resulting textual inputs \(\{r, y_i\}\) are encoded as \(\mathbf{w}_i = \mathrm{E}_\mathrm{T}(\{r, y_i\}) \in \mathbb{R}^d\), and \(d\) is the embedding dimension.  Given a point cloud \(\mathbf{X} \in \mathbb{R}^{L \times 3}\), its embedding \(\mathbf{f} = \mathrm{E}_\mathrm{P}(\mathbf{X}) \in \mathbb{R}^d\) is compared to \(\mathbf{w}_i\) via cosine similarity, and the probability distribution is given as follows:

\vspace{-0.2cm}

\begin{equation}
p(y_i|\mathbf{X}) = \frac{\exp(\text{sim}(\mathbf{w}_i, \mathbf{f})/\tau)}{\sum_{j=1}^K \exp(\text{sim}(\mathbf{w}_j, \mathbf{f})/\tau)}
\end{equation}
where \(\text{sim}(\cdot, \cdot)\) denotes the cosine similarity and \(\tau\) is the temperature controlling the sharpness of the distribution.

\begin{figure*}
    \centering
    \includegraphics[width=1\textwidth]{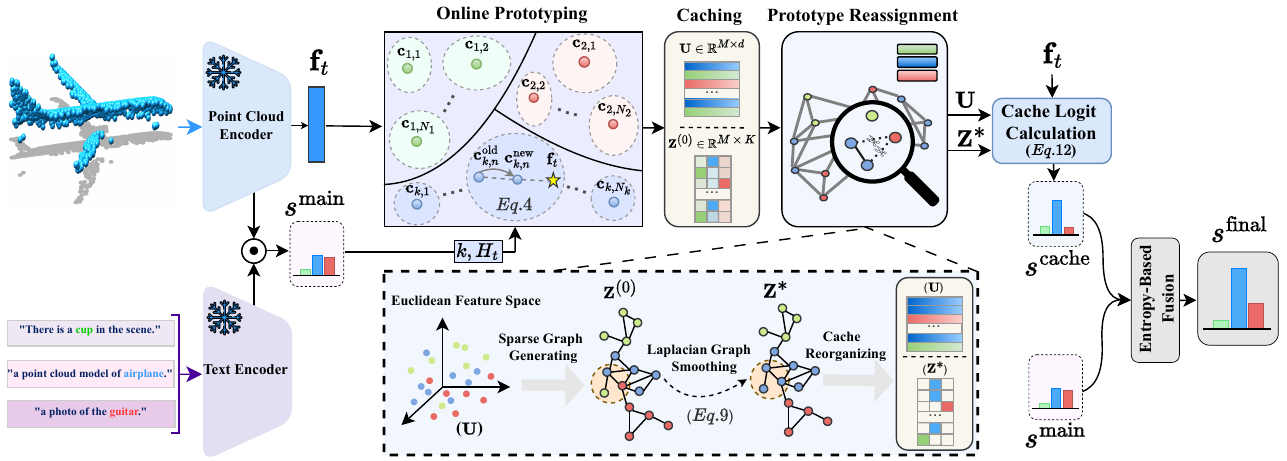}

    \caption{{Method Overview.} Given a test point cloud \( \mathbf{X}_t \in \mathbb{R}^{L \times 3} \), our method extracts a point cloud feature \( \mathbf{f}_t \) via a point cloud encoder. The {3D cache} is updated via {online Prototyping}, where cluster centers serve as {3D prototypes}. The {Prototype Reassignment} module refines these prototypes, and their affinity with \( \mathbf{f}_t \) is computed to obtain \( \mathbf{s}^{\text{cache}} \). Finally, the prediction logit \( \mathbf{s}^{\text{final}} \) is obtained by fusing \( \mathbf{s}^{\text{cache}} \) and the model’s base output \( \mathbf{s}^{\text{main}} \) using entropy-driven confidence weighting.
    }

    \vspace{-0.5cm}
    \label{fig:method}
\end{figure*}

\subsection{Uni-Adapter Method}
The overall framework of the proposed method is illustrated in Fig.~\ref{fig:method}. It integrates similarity scores from point cloud-to-text comparisons with those from the cache model. The cache model learns 3D prototypes using an \textit{online prototyping} module and dynamically refines noisy prototype assignment through \textit{prototype reassignment} module. It then computes cache scores based on the affinity between the input point cloud representation and the stored prototypes. Finally, these scores are unified based on the entropy of the predictions to obtain the final similarity score. The following sections provide a detailed description of each component.

\subsubsection{Online Prototyping Module.}
We adopt an \textit{online clustering} strategy, termed \textit{online prototyping}, to dynamically capture diverse modes of the data distribution. This module incrementally updates a set of class-specific prototypes. The goal is to associate each incoming point cloud feature with a representative prototype and update it accordingly. 

At time step \( t \), a point cloud \( \mathbf{X}_t \) is encoded as
\(
  \mathbf{f}_t = \mathrm{E}_{\mathrm{P}}(\mathbf{X}_t) \in \mathbb{R}^d.  
\)
\noindent We first predict the class \(k\) by computing cosine similarities between \( \mathbf{f}_t \) and class embeddings \( \{\mathbf{w}_i\}_{i=1}^K \):
\vspace{-0.1cm}
\begin{equation}
    s_i^{\text{main}} = \mathrm{sim}(\mathbf{w}_i, \mathbf{f}_t), \quad
k = \arg\max_i s_i^{\text{main}}.
\end{equation}
\noindent Each class \( k \) maintains up to \( N \) prototypes, 
denoted by \( \{\mathbf{c}_{k,j} \in \mathbb{R}^d\}_{j=1}^{N_k} \), 
where \( N_k \leq N \).  
Given the predicted class \( k \), we select the most similar prototype:
\vspace{-0.1cm}
\begin{equation}
    n = \arg\max_{1 \leq j \leq N_k} \mathrm{sim}(\mathbf{f}_t, \mathbf{c}_{k,j}).
\end{equation}

\noindent If an empty slot exists (\( N_k < N \)), it is initialized with \( \mathbf{f}_t \). Otherwise, the selected prototype \( \mathbf{c}_{k,n} \) is updated using a confidence-weighted moving average:
\vspace{-0.1cm}
\begin{equation} 
\mathbf{c}_{k,n}^{\text{new}} = \frac{\alpha_t \mathbf{f}_t + b_{k,n} \alpha_{k,n} \mathbf{c}_{k,n}^{\text{old}}}{\alpha_t + b_{k,n} \alpha_{k,n}},
\end{equation}

\noindent where \( b_{k,n} \) is the number of past updates to the prototype, and \( \alpha_t \), \( \alpha_{k,n} \) are the confidence scores of the incoming sample and the cached prototype, respectively. These scores are derived from prediction entropy as:
\begin{equation}
\alpha_t = \exp(-\beta \cdot H_t), \quad \alpha_{k,n} = \exp(-\beta \cdot H_{k,n}),
\end{equation}
\vspace{-0.2cm}

\noindent where \( \beta \) is a scaling factor, and \( H_t \) and \( H_{k,n} \) denote the entropy of the softmax over similarities with text embeddings. Specifically, \( H_t \) is computed from feature \( \mathbf{f}_t \), and \( H_{k,n} \) from prototype \( \mathbf{c}_{k,n} \), both compared against \( \{ \mathbf{w}_i \}_{i=1}^{K} \).

\subsubsection{Prototype Reassignment Module.}

While online prototyping maintains representative prototypes, it remains sensitive to noisy pseudo-labels. To improve label reliability, we introduce a \textit{prototype reassignment} module that smooths pseudo-labels across similar prototypes via graph-based regularization. To refine pseudo-labels based on semantic relationships, we require two components: (1) a similarity matrix capturing prototype relationships, and (2) the initial soft pseudo-labels to be updated. These soft pseudo-labels, given by the model’s softmax probabilities over class logits, are stored in \( \mathbf{Z}^{(0)} \in \mathbb{R}^{M \times K} \), where each row corresponds to a prototype and contains its class probabilities. 

Let \( M = \sum_{k=1}^K N_k \) denote the total number of active prototypes across all classes, where \( N_k \leq N \) is the number of prototypes for class \( k \). We collect all prototype features into a matrix \( \mathbf{U} = [\mathbf{c}_{1,1}; \dots; \mathbf{c}_{K,N_K}] \in \mathbb{R}^{M \times d} \), where each row is a \( \ell_2 \)-normalized prototype. The similarity matrix is computed as:
\vspace{-0.1cm}
\begin{equation}
\mathbf{A} = \mathbf{U} \mathbf{U}^\top \in \mathbb{R}^{M \times M}.
\end{equation}

We apply a threshold \( \gamma \in [0,1] \) to remove weak connections and obtain a sparse matrix \( \hat{\mathbf{A}} \) by zeroing out values below \( \gamma \). From \( \hat{\mathbf{A}} \), we compute the degree matrix \( \mathbf{D} \), a diagonal matrix where each diagonal entry \( \mathbf{D}_{mm} \) is the sum of the \( m \)-th row in \( \hat{\mathbf{A}} \). The normalized graph Laplacian is then:
\begin{equation}
\mathbf{L}_{\text{norm}} = \mathbf{I} - \mathbf{D}^{-1/2} \hat{\mathbf{A}} \mathbf{D}^{-1/2}.
\end{equation}

\noindent This refinement is formulated as the following optimization:
\vspace{-0.2cm}
\begin{equation}
\mathbf{Z}^* = \arg\min_{\mathbf{Z}} \| \mathbf{Z} - \mathbf{Z}^{(0)} \|_F^2
+ \lambda_{\text{reg}} \cdot \text{Tr}(\mathbf{Z}^\top \mathbf{L}_{\text{norm}} \mathbf{Z}),
\end{equation}
where \( \lambda_{\text{reg}} > 0 \) balances fidelity to the initial predictions and label smoothness across the graph. This objective has a closed-form solution:
\begin{equation}
\mathbf{Z}^* = \left( \mathbf{I} + \lambda_{\text{reg}} \mathbf{L}_{\text{norm}} \right)^{-1} \mathbf{Z}^{(0)}.
\label{eq:inverse}
\end{equation}
Finally, we convert \( \mathbf{Z}^* \) into a one-hot label matrix by keeping the maximum entry in each row:

\vspace{-0.3cm}
\begin{equation}
\mathbf{Z}^*_{m,\hat{i}} =
\begin{cases}
1 & \text{if } \hat{i} = \arg\max_{i} \mathbf{Z}^*_{m,i}, \\
0 & \text{otherwise}
\end{cases}
\quad \text{for } m = 1, \ldots, M.
\end{equation}

It should be noted that, to reduce computational overhead (\( \mathcal{O}(M^3) \)), we solve Equation \ref{eq:inverse} using the conjugate gradient method~\cite{hestenes1952methods}, which reduces the complexity to \( \mathcal{O}(\rho \cdot \text{nnz}(\mathbf{L}_{\text{norm}})) \), where \( \rho \) is the number of iterations and \( \text{nnz}(\cdot) \) denotes the number of non-zero entries.

\subsubsection{Cache Logit Calculation.}

Each prototype now has a one-hot class label encoded in \( \mathbf{Z}^* \in \{0,1\}^{M \times K} \). Given an input feature \( \mathbf{f}_t \in \mathbb{R}^d \), we compute its cosine similarity to all prototypes as: \(\mathbf{U} \mathbf{f}_t \in \mathbb{R}^M.
\)
To ensure that the similarity scores are not biased by the number of prototypes per class, we normalize by the count of prototypes assigned to each class. Specifically, we compute a diagonal normalization matrix:
\begin{equation}
\boldsymbol{\Lambda} = \text{diag} \left( \left( \frac{1}{\sum_{m=1}^{M} \mathbf{Z}^*_{m,i}} \right)_{i=1}^{K} \right) \in \mathbb{R}^{K \times K},
\end{equation}
where the \( i \)-th diagonal entry rescales the summed similarity for class \( i \) by the number of associated prototypes. The cache-based logits are then computed as:
\begin{equation}
\mathbf{s}^{\text{cache}} = \boldsymbol{\Lambda} \, \mathbf{Z}^{*\top} \left( \mathbf{U} \mathbf{f}_t \right) \in \mathbb{R}^K.
\end{equation}

This yields class-wise average similarities between \( \mathbf{f}_t \) and the prototypes. The resulting \( \mathbf{s}^{\text{cache}} \) is fused with main logits for robust classification.

\begin{table*}[t!]
    \centering

    \setlength{\tabcolsep}{4pt}
    \renewcommand{\arraystretch}{1.3}
    \resizebox{0.9\linewidth}{!}{
    \begin{tabular}{cl|ccccccccccccccc|c}

    \Xhline{3\arrayrulewidth}

    & \bf Method & \rotatebox{60}{uni} & \rotatebox{60}{gauss} & \rotatebox{60}{backg} & \rotatebox{60}{impul} & \rotatebox{60}{upsam} & \rotatebox{60}{rbf} & \rotatebox{60}{rbf-inv} & \rotatebox{60}{den-dec} & \rotatebox{60}{dens-inc} & \rotatebox{60}{shear} & \rotatebox{60}{rot} & \rotatebox{60}{cut} & \rotatebox{60}{distort} & \rotatebox{60}{oclsion} & \rotatebox{60}{lidar} &\bf Avg.  \\ \Xhline{2\arrayrulewidth}

& Source-Only  &  {57.37} &             {54.01} &              {70.21} &             {61.91} &             {60.69} &             {51.74} &             {52.39} &             {67.50} &             {74.87} &             {72.40} &             {71.02} &             {63.97} &             {58.95} &             {47.24} &             {22.93} &             {59.15}  \\ \cdashline{2-18}

\multirow{5}{*}{\rotatebox{90}{\textbf{Training}}} 
& TENT (ICLR21)    &    {61.46} &             {58.30} &             {65.28} &             {56.36} &             {65.08} &             {51.62} &             {52.51} &             {65.44} &             {75.24} &             {72.36} &             {72.44} &             {61.43} &             {58.18} &             {48.09} &             {28.44} &             {59.48} \\

& SHOT (ICML20)       &      {61.50} &             {58.42} &              {65.27} &             {56.68} &             {64.66} &             {51.54} &             {52.84} &             {65.48} &             {75.57} &             {72.81} &             {72.29} &             {61.14} &             {58.14} &             {48.46} &             {28.44} &             {59.55} \\

& SAR (ICLR23) &     {61.51} &             {58.18} &             {65.40} &             {56.77} &             {64.79} &             {51.86} &             {53.40} &             {66.37} &             {75.93} &             {72.49} &             {72.08} &             {62.28} &             {59.12} &             {49.31} &             {30.71} &             {60.01} \\    
& DUA (CVPR22)      &   {61.26} &             {57.90} &             {65.11} &             {56.76} &             {64.79} &             {51.38} &             {53.08} &             {65.44} &             {75.16} &             {72.93} & {72.41} &             {60.85} &             {58.35} & {48.50} & {28.65} &             {59.50} \\     & MEMO (NIPS22)&      {57.38} &             {54.08} &             70.24 &             {61.92} &             {60.71} &             {51.76} &             {52.39} &             {67.59} &             {74.88} &             {72.48} &             {71.03} &             {63.98} &             {59.30} &             {47.36} &             {22.94} &             {59.20} \\

 & TPT* (NIPS22) &   {60.65} &             {57.20} &             \underline{76.32} &             {61.15} &             {63.47} &             {55.39} &             {55.78} &             {71.36} &             {74.77} &             {75.20} &             {73.01} &             {65.41} &             {60.97} &             {47.60} &             {18.10} &             {61.02}


\\ \cmidrule(lr){2-18}

 & LAME (CVPR22) &             {57.33} &             {54.09} &  {70.10} &             {61.63} &             {60.94} &             {51.94} &             {52.27} &             {67.50} &             {75.16} &             {72.45} &             {71.27} &             {63.90} &             {59.04} &             {47.33} &             {22.61} &             {59.17} \\

\multirow{3}{*}{\rotatebox{90}{\textbf{Training-Free}}}
& 3DD-TTA (WACV25) &             {60.53} &             {59.76} &              {63.53} &             {64.71} &             {67.91} &             {50.73} &             {51.09} &             {59.36} &             {66.98} &             {64.26} &             {58.67} &             {59.36} &             {53.77} &             {36.10} &             32.37 &             {56.06} \\

& CloudFixer $\dagger$  (ECCV24) &             {65.44} &             \underline{65.84} &              {63.90} &             68.11 &             \bfseries{72.77} &             {52.80} &             {53.81} &             {52.88} &             {63.21} &             {59.36} &             {61.18} &             {56.16} &             {52.18} &             {29.09} &             {24.60} &             {56.09} \\

& T3A (NIPS21)&      \bfseries{69.40} &             \bfseries{70.26} & {41.89} &             63.33 &             {70.74} &             61.26 &             59.27 &             72.20 &             \underline{79.33} &             \underline{77.59} &             \underline{78.36} &             \underline{71.07} &             65.51 &         49.59 &              32.01 &             64.12 \\

& TDA* (CVPR24) & {62.20} & {61.63} & {75.16} & {65.52} & {67.18} & {57.46} & {57.94} & {70.95} & {76.94} & {74.43} & {73.26} & {67.46}& {63.17} & {50.69} & {30.39} & {63.63} \\

& Point-Cache* (CVPR25) & 64.34 & {64.87} & {73.95} & \underline{68.31} & {71.68} & \underline{62.84} & \underline{65.19} & \underline{73.22} & 77.80 & 77.15 & {75.77} & {69.77} & {\underline{68.31}} & {\underline{54.78}} & \underline{32.98} & \underline{66.73} \\

 \cmidrule(lr){2-18}
    & \ccol \bf Uni-Adapter (Ours) & \ccol \underline{66.82} &  \ccol  {65.52} &         \ccol    \bfseries{78.32} &  \ccol \bfseries{72.25} &  \ccol \underline{72.04} &  \ccol  \bfseries{65.60} &  \ccol \bfseries{66.61} & \ccol  \bfseries{77.51} &  \ccol \bfseries{80.63} &  \ccol \bfseries{79.05}&  \ccol \bfseries{79.30} &  \ccol \bfseries{75.29} &  \ccol \bfseries{73.38} &  \ccol \bfseries{56.92} &  \ccol 
 \bfseries{36.26} &  \ccol \bfseries{69.70} \\
         \bottomrule
    \Xhline{3\arrayrulewidth}
\end{tabular}}
\caption{Top-1 accuracy (\%) on ModelNet40-C under distribution shifts using Uni3D-Large (batch size = 1). Source-Only shows performance without adaptation. Best and second-best are in \textbf{bold} and \underline{underline}. \textbf{*} denotes VLFM-based TTA methods.
}
\label{tab:1_modelnet_c}
\vspace{-0.25cm}
\end{table*}

\subsubsection{Entropy-Based Fusion.}

We combine the predictions from the source VLFM and the cache model by fusing their logits. The fusion is performed using entropy-based weighting:
\begin{equation}
\boldsymbol{s}^{\text{final}} = \frac{H_{\text{cache}} \cdot \boldsymbol{s}^{\text{main}} + H_t \cdot \boldsymbol{s}^{\text{cache}}}{H_{\text{cache}} + H_t}.
\end{equation}

\noindent Here, \( H_t \) and \( H_{\text{cache}} \) are the entropies of the softmax over the main and cache logits, respectively. The fusion adaptively weights each source by its confidence, favoring the more certain modality.

\section{Experiments}
\label{sec:experiments}


\begin{table*}[t!]
    \centering

    \setlength{\tabcolsep}{4pt}
    \renewcommand{\arraystretch}{1.3}
    \resizebox{0.9\linewidth}{!}{
    \begin{tabular}{cl|ccccccccccccccc|c}

    \Xhline{3\arrayrulewidth}

    & \bf Method & \rotatebox{60}{uni} & \rotatebox{60}{gauss} & \rotatebox{60}{backg} & \rotatebox{60}{impul} & \rotatebox{60}{upsam} & \rotatebox{60}{rbf} & \rotatebox{60}{rbf-inv} & \rotatebox{60}{den-dec} & \rotatebox{60}{dens-inc} & \rotatebox{60}{shear} & \rotatebox{60}{rot} & \rotatebox{60}{cut} & \rotatebox{60}{distort} & \rotatebox{60}{oclsion} & \rotatebox{60}{lidar} &\bf Avg.  \\ \Xhline{2\arrayrulewidth}

& Source-Only & {60.33} & {55.75} & {65.95} & {65.02} & {59.04} & {59.41} & {60.23} & {79.06} & {71.07} & {75.62} & {73.87} & {76.82} & {63.22} & {2.37} & {1.06} & {57.92} \\ \cdashline{2-18}

\multirow{5}{*}{\rotatebox{90}{\textbf{Training}}}

& TENT (ICLR21)       & {59.88} & {54.30} & {58.31} & {61.14} & {57.78} & {58.69} & {60.17} & {79.30} & {72.69} & {75.83} & {74.54} & {77.08} & {63.23} & {2.97} & \bfseries{2.54} & {57.23} \\

& SHOT (ICML20)    & {59.96} & {54.32} & {58.37} & {61.14} & {57.67} & {58.81} & {60.19} & {79.23} & {72.53} & {75.90} & {74.57} & {77.09} & {63.34} & {3.02} & {2.51} & {57.24} \\

& SAR (ICLR23) & {59.86} & {54.09} & {58.59} & {60.84} & {57.48} & {58.65} & {60.09} & {79.04} & {73.23} & {74.78} & {71.39} & {77.04} & {61.88} & {2.69} & {1.40} & {56.74} \\    
& DUA (CVPR22) & {59.85} & {54.37} & {58.26} & {61.21} & {57.89} & {58.71} & {60.15} & {79.11} & {72.40} & {72.55} & \underline{75.91} & {77.05} & {63.24} & {2.97} & \underline{2.53} & {57.08} \\

& MEMO (NIPS22) & {60.33} & {55.76} & {66.02} & {65.02} & {59.04} & {59.42} & {60.23} & {79.01} & {70.92} & {75.62} & {73.95} & {76.82} & {63.22} & {2.41} & {1.09} & {57.92} \\

 & TPT* (NIPS22) &      {62.87} &             {56.63} &             \underline{69.20} &             {64.70} &             {59.10} &             {58.29} &             {60.43} &             \bfseries{81.59} &             \underline{75.23} &             \underline{76.93} &             {74.56} &             \bfseries{80.52} &             {63.02} &             {2.17} &             {1.25} &             {59.10}

\\ \cmidrule(lr){2-18}

 & LAME (CVPR22) & {60.43} & {55.89} & 66.04 & {65.12} & {59.09} & {59.42} & {60.33} & {79.13} & {71.16} & {75.74} & {74.08} & {76.99} & {63.35} & {2.31} & {1.02} & {58.01} \\
 \multirow{3}{*}{\rotatebox{90}{\textbf{Training-Free}}}   
& 3DD-TTA (WACV25)& \underline{65.78} & \underline{64.16} & {55.00} & {61.75} & \bfseries{68.05} & {55.06} & {56.20} & {74.20} & {67.35} & {68.06} & {61.19} & {73.01} & {56.29} & {2.50} & {0.97} & {55.30} \\

& CloudFixer $\dagger$  (ECCV24) &  65.57 & \bfseries{65.30} & {58.15} & \bfseries{69.53} & {63.65} & {55.02} & {56.71} & {69.89} & {58.67} & {65.65} & {65.64} & {70.36} & {55.09} & {3.75} & {2.46} & {57.24} \\

& T3A (NIPS21) & 60.60 & {53.12} & {20.70} & {44.19} & {49.31} & {46.35} & {44.37} & {70.31} & {63.01} & {64.86} & {63.83} & {68.24} & {52.60} & 1.00 & {0.87} & {46.89} \\

& TDA* (CVPR24) & {62.75} & {58.95} & {68.33} & {67.14} & {62.09} & \underline{61.28} & \underline{62.56} & {79.00} & {71.44} & {75.93} & {74.79}& {77.17} & \underline{64.44} & \underline{3.82} & {1.81} & \underline{59.43}\\

& Point-Cache* (CVPR25) & 62.63 & {56.71} & 66.51 & 65.85 & 61.15 & {59.79} & {61.49} & 75.89 & 69.47 & 72.61 & 70.81 & 73.82 & 63.41 & 3.64 & 1.67 & 57.70 \\

 \cmidrule(lr){2-18}
    & \ccol \bf Uni-Adapter (Ours) & \ccol \bfseries{66.89} &  \ccol  62.23 &         \ccol    \bfseries{71.38} &  \ccol \underline{68.62} &  \ccol \underline{64.15} &  \ccol  \bfseries{67.42} &  \ccol \bfseries{67.33} & \ccol  \underline{80.76} &  \ccol \bfseries{75.69} &  \ccol \bfseries{78.11}&  \ccol \bfseries{77.01} &  \ccol \underline{79.64} &  \ccol \bfseries{70.14} &  \ccol \bfseries{4.36} &  \ccol 
 {2.47} &  \ccol \bfseries{62.41 } \\
         \bottomrule
    \Xhline{3\arrayrulewidth}
\end{tabular}}
\caption{Top-1 accuracy (\%) on ShapeNet-C under distribution shifts using Uni3D-Large (batch size = 1). Source-Only shows performance without adaptation. Best and second-best are in \textbf{bold} and \underline{underline}. \textbf{*} denotes VLFM-based TTA methods.
}
\label{tab:2_shapenet_c}
\vspace{-.3cm}
\end{table*}


\begin{table*}[t!]
    \centering

    \label{tab:2}
    \setlength{\tabcolsep}{4pt}
    \renewcommand{\arraystretch}{1.3}
    \resizebox{0.9\linewidth}{!}{
    \begin{tabular}{cl|ccccccccccccccc|c}

    \Xhline{3\arrayrulewidth}

    & \bf Method & \rotatebox{60}{uni} & \rotatebox{60}{gauss} & \rotatebox{60}{backg} & \rotatebox{60}{impul} & \rotatebox{60}{upsam} & \rotatebox{60}{rbf} & \rotatebox{60}{rbf-inv} & \rotatebox{60}{den-dec} & \rotatebox{60}{dens-inc} & \rotatebox{60}{shear} & \rotatebox{60}{rot} & \rotatebox{60}{cut} & \rotatebox{60}{distort} & \rotatebox{60}{oclsion} & \rotatebox{60}{lidar} &\bf Avg.  \\ \Xhline{2\arrayrulewidth}

& Source-Only & {29.78} & {25.99} & {40.62} & {45.96} & {30.64} & {33.05} & {34.42} & {56.28} & {47.16} & {54.22} & {54.04} & {56.80} & {43.55} & {9.98} & {8.61} & {38.07} \\ \cdashline{2-18}

\multirow{5}{*}{\rotatebox{90}{\textbf{Training}}} 

& TENT (ICLR21)     & {29.78} & {26.16} & {49.91} & {51.46} & {30.65} & {33.22} & {36.32} & {55.08} & {45.27} & {52.50} & {53.18} & {55.77} & {44.92} & {7.06} & {3.79} & {38.34} \\

& SHOT (ICML20)    & {29.60} & {26.85} & {51.12} & 52.32 & {31.33} & {33.73} & {37.20} & {56.80} & {45.78} & {54.57} & {54.22} & {56.31} & {45.27} & {6.72} & {3.96} & {39.05} \\

& SAR (ICLR23) & {29.08} & {27.54} & {42.17} & {44.58} & {31.33} & {32.36} & {34.77} & {56.28} & {44.41} & {52.84} & {54.22} & {55.59} & {44.06} & {9.64} & \underline{9.64} & {37.90} \\    
& DUA (CVPR22) & {29.95} & {27.37} & {41.65} & {44.92} & {30.81} & {32.53} & {34.08} & {56.63} & {45.09} & {53.87} & {54.22} & {55.77} & {43.89} & {9.81} & {9.47} & {38.00} \\

& MEMO (NIPS22)& {29.77} & {26.16} & {49.91} & {51.46} & {30.64} & {33.22} & {36.32} & {55.08} & {45.27} & {52.50} & {53.18} & {55.77} & {44.92} & {7.05} & {3.78} & {38.34} \\  

 & TPT* (NIPS22) &      {30.04} &             {28.43} &             40.95 &             {46.76} &             {32.68} &             {35.55} &             {34.39} &             {56.60} &             {50.66} &             {53.81} &             {54.39} &             {60.30} &             {42.70} &             \underline{11.13} &             {6.07} &             {38.96}

\\ \cmidrule(lr){2-18}

 & LAME (CVPR22) & {29.60} & {26.85} & 51.12 & {52.32} & {31.33} & {33.73} & {37.18} & {56.80} & {45.78} & {54.56} & {54.22} & {56.29} & {45.27} & {6.71} & {3.96} & {39.05} \\
\multirow{3}{*}{\rotatebox{90}{\textbf{Training-Free}}}    
& 3DD-TTA (WACV25) & {32.19} & {30.81} & {27.71} & {39.59} & {34.60} & {25.82} & {26.51} & {45.61} & {38.04} & {36.14} & {33.39} & {45.09} & {33.05} & {8.61} & {4.47} & {30.78} \\

& CloudFixer $\dagger$  (ECCV24) & \bfseries{36.83} & \underline{33.22} & {36.14} & {48.19} & \underline{37.69} & {27.54} & {30.46} & {45.44} & {40.28} & {38.73} & {38.21} & {46.13} & {35.28} & {10.67} & \bfseries{11.70} & {34.43} \\

& T3A (NIPS21) & {34.94} & {32.70} & {34.08} & {43.37} & {33.22} & {36.49} & 36.32 & \bfseries{62.65} & \bfseries{55.42} & \underline{61.46} & {\bfseries{62.31}} & {\bfseries{64.03}} & {\bfseries{56.97}} & {{9.47}} & {8.61} & \underline{42.14} \\

& TDA* (CVPR24) & {31.33} & {28.40} & {52.67} & {53.36} & {32.36} & {36.32} & {40.45} & {56.80} & {46.82} & {55.25} & {55.76}& {57.66} & {49.40} & {8.09} & {4.65} & {40.62}\\

& Point-Cache* (CVPR25) & 30.98 & {27.19} & \bfseries{55.25} & \underline{56.45} & 33.73 & \underline{40.79} & \underline{43.03} & 59.50 & 49.23 & 60.07 & {56.63} & {57.66} & {49.05} & {8.26} & 4.30 & 42.13 \\

 \cmidrule(lr){2-18}
    & \ccol \bf Uni-Adapter (Ours) & \ccol \underline{35.28} &  \ccol  \bfseries{37.69} &         \ccol    \underline{53.35} &  \ccol \bfseries{59.55} &  \ccol \bfseries{39.07} &  \ccol  \bfseries{42.16} &  \ccol \bfseries{52.49} & \ccol  \underline{60.58} &  \ccol \underline{51.97} &  \ccol \bfseries{61.61}&  \ccol \underline{60.24} &  \ccol \underline{60.92} &  \ccol \underline{54.38} &  \ccol \bfseries{20.82} &  \ccol 
 {4.81} &  \ccol \bfseries{ 46.33} \\
         \bottomrule
    \Xhline{3\arrayrulewidth}
\end{tabular}}
\caption{Top-1 accuracy (\%) on ScanObjectNN-C using Uni3D-Large (batch size = 1). \textbf{*} denotes VLFM-based TTA. Source-Only shows performance without adaptation. Best and second-best are in \textbf{bold} and \underline{underline}.
}
\label{tab:3_scanobjecnn_c}
\end{table*}

\begin{table}[!t]
   \small 
   \centering
   \resizebox{\linewidth}{!}{ 
   \begin{tabular}{l c c c c c}
      \toprule
      \multirow{2}{*}{\textbf{Method}} 
      & \multicolumn{3}{c}{\textbf{Small-Scale Data }} 
      & \multicolumn{2}{c}{\textbf{Large-Scale Data }} \\
      \cmidrule(lr){2-4} \cmidrule(lr){5-6}
            & ModelNet & SONN & ShapeNet 
            & O-LVIS & Omni3D \\
      \midrule
      ULIP-2 (CVPR24) & 71.23 & 52.49 & 69.53  & 30.26  & 26.36  \\ 
      \ +Point-Cache (CVPR25) & 74.53 & 58.52 & 69.74 & \textbf{32.36} & 29.38 \\
      \ \textbf{+Uni-Adapter (Ours)} & \ccol{\textbf{76.47}} & \ccol{\textbf{58.84}} & \ccol{\textbf{70.81}} & \ccol{31.83} & \ccol{\textbf{30.37}} \\ 
      \midrule
      O-Shape (NIPS23) & 84.56 & 55.94 & 74.53 & 46.15 & 34.09 \\
      \ +Point-Cache (CVPR25) & 84.04 & 62.48 & 78.51 & 46.05 & 34.45 \\
      \ \textbf{+Uni-Adapter (Ours)} & \ccol{\textbf{85.44}} & \ccol{\textbf{63.12}} & \ccol{\textbf{79.99}} & \ccol{\textbf{48.07}} & \ccol{\textbf{37.21}} \\ 
      \midrule
      Uni3D (ICLR24) & 78.84 & 59.55 & 79.90 & 46.20 & 32.06 \\
      \ +Point-Cache (CVPR25) & 83.43 & 62.27 & 78.51 & 47.13 & 33.22 \\
      \ \textbf{+Uni-Adapter (Ours)} & \ccol{\textbf{83.96}} & \ccol{\textbf{64.03}} & \ccol{\textbf{81.23}} & \ccol{\textbf{47.49}} & \ccol{\textbf{35.95}} \\ 
      \bottomrule
   \end{tabular}
   } 
    \caption{Performance of Uni-Adapter on 3D VLFMs across clean datasets (batch size = 1). Objaverse-LVIS uses 10,000 points; others use 1,024. SONN denotes ScanObjectNN.}
   \label{tab4:recognition_accuracy}
   \vspace{-0.2cm}
\end{table}

%% file: sec/4_experiments.tex
\subsection{Experimental Setup}
\subsubsection{Datasets.}
We evaluate our approach under distribution shifts using ModelNet-40C \cite{modelnet40c}, ShapeNet-C \cite{mirza2022mate}, and ScanObjectNN-C \cite{mirza2022mate}, which introduce 15 types of synthetic corruptions, including density variations, noise, and geometric transformations, each with five severity levels. To further assess the generalization capability on unseen data, we conduct experiments on the test splits of ModelNet40 \cite{wu15modelnet}, ShapeNetCore-v2 \cite{chang2015shapenet}, and ScanObjectNN \cite{uy2019revisiting}, as well as large-scale 3D datasets such as OmniObject3D \cite{OmniObject3D} (216 classes) and Objaverse-LVIS \cite{Objaverse} (1,156 classes), designed to evaluate generalization across diverse categories.

\vspace{0.1cm}
\noindent\textbf{Baselines.}
To evaluate our Uni-Adapter and ensure a fair comparison, we implemented twelve diverse baselines spanning both training-free and training-based TTA approaches. Specifically, we evaluate  TENT \cite{Tent}, SHOT \cite{shot}, SAR \cite{sar}, DUA \cite{dua}, MEMO \cite{memo}, LAME \cite{lame}, T3A \cite{T3A}, CloudFixer \cite{shim2024cloudfixer}, 3DD-TTA \cite{dastmalchi2024test}, TPT \cite{tpt}, TDA \cite{TDA}, and Point-Cache \cite{Point-Cache}. While CloudFixer, 3DD-TTA, and Point-Cache are designed for 3D point clouds, the remaining methods originate from the 2D domain and are adapted for 3D data. For CloudFixer, we use only its generative model and guidance, without updating the source model, denoted as CloudFixer~$\dagger$. Note that TPT and TDA are developed specifically for 2D VLFMs, whereas Point-Cache is natively built for 3D VLFMs.

\vspace{0.1cm}
\subsubsection{Implementation.}
We use ULIP-2~\cite{ulip2}, OpenShape~\cite{openshape}, and Uni3D-Large~\cite{zhou2023uni3d} as 3D VLFMs. Test-time adaptation is performed on a single sample. For Graph-Based Label Smoothing, we set the sparsity threshold \( \gamma = 0.5 \) to retain strong correlations in the adjacency matrix and the confidence decay parameter \( \beta = 10 \) to balance diversity and confidence when updating cluster centers.
Each target sample has 1024 points, except Objaverse-LVIS with 10{,}000. All experiments use corruption severity level 5 on a single NVIDIA RTX 4090 GPU.

\vspace{-0.1cm}
\subsection{Results}
\label{subsec:results}

\subsubsection{Robustness Against Distribution Shifts.} We evaluate Uni-Adapter on ModelNet-40C (Table~\ref{tab:1_modelnet_c}) and ShapeNet-C (Table~\ref{tab:2_shapenet_c}) across diverse corruption types. While most training-based baselines adapted from the 2D domain show only marginal gains, methods such as T3A, TDA, and Point-Cache yield relatively stronger improvements. We also observe that, although 3D input adaptation methods \cite{shim2024cloudfixer, dastmalchi2024test}  address specific types of corruption, their deployment on novel test instances may introduce excessive generation errors and greater deviations from the source domain, ultimately amplifying distribution shift. Moreover, methods designed for VLFMs demonstrate stronger capabilities in reducing modality gaps. While test-time methods like T3A, TDA, and Point-Cache outperform traditional 2D baselines on ModelNet-40C, Uni-Adapter surpasses all alternatives, particularly under significant domain shifts, with improvements reaching 10.55\%. On the more diverse ShapeNet-C benchmark, characterized by higher intra-class variance, methods relying solely on confident samples (e.g., T3A, Point-Cache)  lead to suboptimal generalization or performance degradation. In contrast, Uni-Adapter, by integrating both confidence and diversity, effectively mitigates these challenges and enhances the source model performance by 4.49\%.

\subsubsection{Effectiveness Under Severe Distribution Shifts.}

To further evaluate the robustness of our Uni-Adapter, we conduct experiments on ScanObjectNN-C, a challenging variant of ScanObjectNN. ScanObjectNN consists of real-world 3D scans that often contain background clutter and partial observations. ScanObjectNN-C introduces additional  corruptions to simulate more severe real-world disturbances. As shown in Table~\ref{tab:3_scanobjecnn_c}, our method outperforms the source model by 8.26\%, showing strong generalization to significant distribution shifts and robustness to real-world 3D corruptions.

\begin{table}[!b]
   \footnotesize
   \centering
   \begin{tabular}{l c c c}
      \toprule
      Method & Uni3D &  OpenShape &  ULIP-2 \\
      \midrule
      Zero-shot &  \textbf{39.19}  & \textbf{15.90} & \textbf{23.94} \\
      TDA  &  36.02  & 14.43 & 21.78 \\
      Point-Cache &  9.73  & 9.74 & 11.11 \\
      \ccol{\textbf{Uni-Adapter (Ours)}} & \ccol{\underline{36.93}}  & \ccol{\underline{15.06}}  &\ccol{\underline{22.67}}\\
         \bottomrule
   \end{tabular}
\caption{Throughput (\(t/s\)) comparison of 3D VLFMs and cache baselines on ModelNet40-C (batch size = 1). Results are averaged over test samples on an RTX 4090 GPU.}

   \label{tab:efficiency_comparison_ensembel_dataset}
\end{table}

\subsubsection{Generalization.} Uni-Adapter demonstrates superior generalization on unseen, uncorrupted datasets of varying scales (see Table~\ref{tab4:recognition_accuracy}). On three small-scale datasets—ModelNet, ScanObjectNN, and ShapeNet—Uni-Adapter outperforms Point-Cache, improving performance across all three source 3D VLFMs and establishing new state-of-the-art baselines. On large-scale 3D benchmarks, including OmniObject3D and Objaverse-LVIS, which feature a diverse and realistic spectrum of object categories, Uni-Adapter consistently yields absolute performance gains.

\subsubsection{Inference Efficiency.}
We evaluate the inference throughput of Uni-Adapter on the ModelNet40 dataset, where throughput (t/s) refers to the number of test instances processed per second. Table 5 shows that Uni-Adapter incurs a smaller drop in throughput compared to zero-shot inference. This overhead is primarily due to the additional operations involved in online prototyping, prototype Reassignment, and logit computation. These findings highlight that Uni-Adapter is a more efficient approach, achieving substantial accuracy improvements with minimal computational overhead relative to other cache-based baseline methods.

%% file: sec/5_Ablation_Study.tex
\section{Ablation Study}
\label{sec:Ablation_study}

\noindent\textbf{Effectiveness of Different Components of Uni-Adapter:} 
Table \ref{tab:4_ablations} evaluates the effectiveness of Uni-Adapter’s components. Starting with Online Prototyping as the only component (row \#1), we observe a significant improvement over the source model (row \#0) across all benchmarks. In row \#2, adding Prototype Reassignment further reduces the performance gap, improving ModelNet-40C by 1.22\%, with similar trends across other datasets. By leveraging correlated features in the dynamic adjacency graph, this module refines pseudo-labels, mitigates inconsistencies, and dynamically corrects errors. The Prototype Reassignment module introduces negligible overhead by employing a conjugate gradient solver that converges quickly on sparse systems.



\begin{table}[h!]
\centering
\setlength{\tabcolsep}{3pt} 
\renewcommand{\arraystretch}{1.5} 
\resizebox{1.0\columnwidth}{!}{
\begin{tabular}{c c c c c c}
\Xhline{1pt}
\# & \begin{tabular}{c} Online\\ Prototyping \end{tabular}  
& \begin{tabular}{c} Prototype \\ Reassignment \end{tabular}  
& ModelNet-40C & ScanObjectNN-C & ShapeNet-C  \\ 
\midrule
0 & \large \ding{55} & \large \ding{55} & \large 59.15 & \large 38.07 & \large 57.92 \\
1 & \Large \checkmark & \large \ding{55} & \large 68.48 & \large 45.12 & \large 61.79 \\
2 & \Large \checkmark & \Large \checkmark & \large \ccol{\textbf{69.70}} & \large \ccol{\textbf{46.33}} & \large \ccol{\textbf{62.41}} \\
\Xhline{1pt}
\end{tabular}
}
\caption{Ablation of Uni-Adapter components. Metric: top-1 accuracy; \#0 is the source model without adaptation.}
\label{tab:4_ablations}
\end{table}

\begin{figure}[b!]
    \centering
    
    \includegraphics[width=\linewidth]{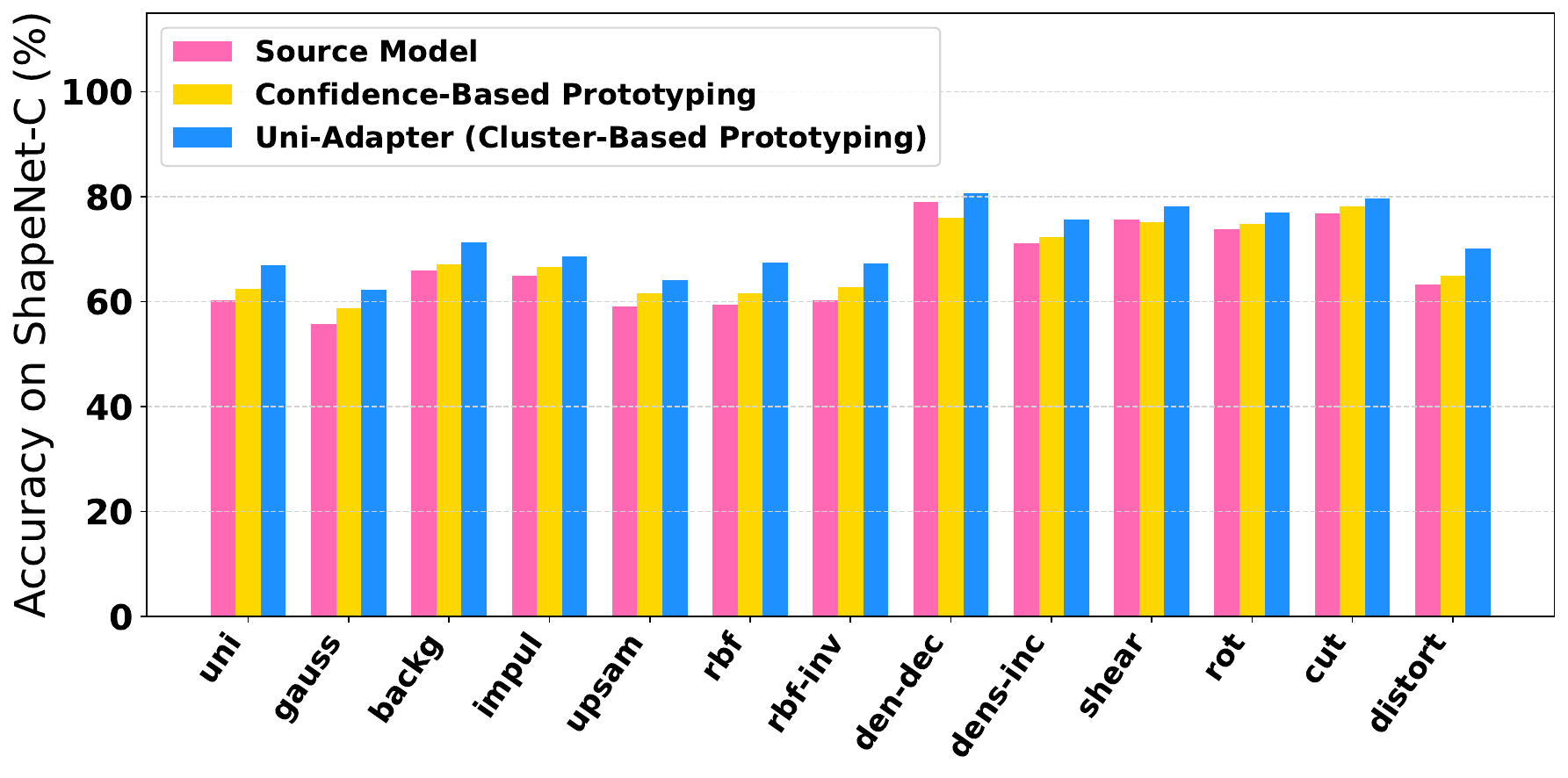} 
    \vspace{-0.6cm}
    \caption{Cluster- vs. confidence-based caches in Uni-Adapter on ShapeNet-C. Cluster-based caching gives higher accuracy by capturing diverse modes, while confidence-based caching misses much of the class distribution.}
    \label{fig:diversity}
\end{figure}

\vspace{0.1cm}
\noindent\textbf{Cluster-Based Cache vs. Confidence-Based Cache:}
We evaluated the effectiveness of our online clustering strategy for learning 3D prototypes by comparing it to a confidence-based cache that retains only the most confident prototypes. Replacing the cluster-based cache in Uni-Adapter with the confidence-based method, we observed that the former consistently outperforms the latter across corruption types in the ShapeNet-C dataset (Fig.~\ref{fig:diversity}).  This improvement demonstrates that online clustering generates more diverse prototypes and better captures the underlying data distribution modes, resulting in a more robust decision boundary.

\begin{figure}
    \centering
    \includegraphics[width=\linewidth]{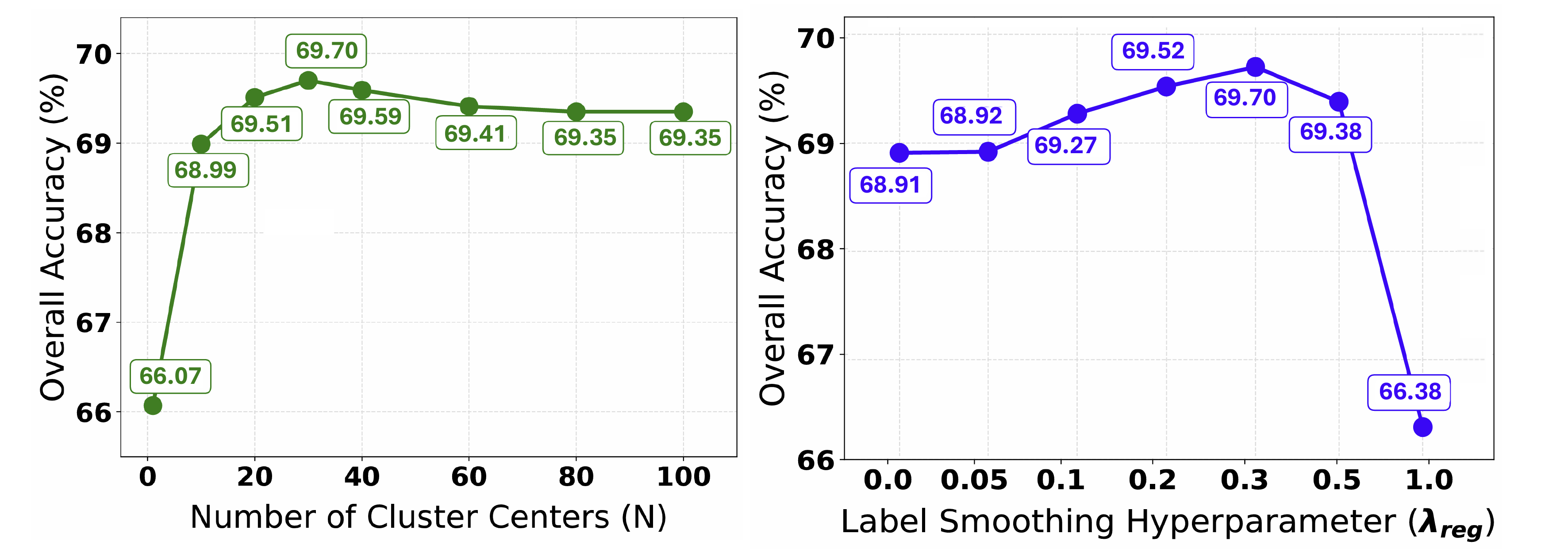}
    \caption{Ablation on the number of cluster centers (N) and label smoothing ($\lambda_{reg}$) for ModelNet-40C.}
    \label{fig:ablation_hyper_parameters}
\end{figure}

\vspace{0.1cm}
\noindent\textbf{Number of Cluster Centers:}  
Fig. \ref{fig:ablation_hyper_parameters} shows the effect of selecting the right number of cluster centers. Too few clusters fail to represent class distributions and limit the effectiveness of Prototype Reassignment module, which requires sufficient features. Too many clusters introduce noise, weakening confidence-based clustering. Our findings suggest that cache size of 30 balances diversity and confidence, with the trade-off controlled by confidence decay hyperparameter \( \beta \).
\vspace{0.1cm}

\noindent\textbf{Graph-based Label Smoothing: } 
We illustrate the impact of the graph-based label smoothing $\lambda_{reg}$ on the performance of the Uni-Adapter (Fig. \ref{fig:ablation_hyper_parameters}). Varying $\lambda_{reg}$ from 0 to 1 controls the refinement intensity. When $\lambda_{\text{reg}}$ is close to 0, the smoothing effect becomes negligible, resulting in minimal pseudo-label refinement and degraded performance. This emphasizes the importance of our Prototype
Reassignment component in enhancing adaptation.  
Conversely, increasing \( \lambda_{reg} \) enhances refinement but can overly smooth labels, diminishing their original influence. We find that \( \lambda_{reg} = 0.3 \) balances effective smoothing with label integrity.

%% file: sec/6_conclusion.tex
\section{Conclusion}

In this paper, we introduce Uni-Adapter, a novel training-free test-time adaptation framework tailored for 3D Vision-Language Foundation Models (VLFMs). Unlike prior confidence-based approaches, Uni-Adapter leverages cluster-based prototypes to capture the multiple mode distribution present in 3D data, enabling more accurate and robust adaptation to real-world variation. By incorporating online prototyping, graph-based prototype reassignment, and entropy-weighted fusion, our method effectively mitigates the challenges of noisy pseudo-labels and preserves semantic consistency across diverse target domains. Extensive experiments across multiple corrupted and real-world 3D benchmarks demonstrate that Uni-Adapter sets a new state of the art in test-time performance for 3D VLFMs, offering a scalable and efficient solution for dynamic and resource-constrained environments. However, Uni-Adapter faces performance instability during the transient cache initialization phase, particularly under severely noisy inputs. Future work may incorporate lightweight self-supervised training using contrastive losses or prototype consistency objectives to improve prototype stability and early-stage adaptation.

%% file: sec/X_suppl.tex
\clearpage
\appendix
\counterwithin{figure}{section}
\counterwithin{table}{section}
\renewcommand\thefigure{\thesection\arabic{figure}}
\renewcommand\thetable{\thesection\arabic{table}}

\centerline{\Large\bf Appendix}

\section{Algorithm of Uni-Adapter} \label{sec:algorithm}  
We present the detailed algorithm for \textit{Uni-Adapter} in Algorithm~\ref{algorithm-uni}, which formalizes our training-free test-time adaptation (TTA) framework for 3D Vision-Language Foundation Models (VLFMs). Uni-Adapter is built around three core components: (1) \textit{online prototyping}, which incrementally maintains a cluster-based cache by performing class-specific online clustering over incoming test features; (2) \textit{prototype assignment}, which addresses pseudo-label noise via graph-based label smoothing over cached prototypes, solved efficiently through the Conjugate Gradient method; and (3) \textit{cache logit calculation}, where affinity-based predictions from the cache are combined with 3D VLFM outputs using an entropy-aware confidence weighting strategy. By maintaining multiple cluster centers per class, Uni-Adapter effectively captures the intrinsic distribution modes within the feature space, enabling more stable and adaptive test-time predictions. The algorithm is modular, lightweight, and scalable, making it highly practical for real-time deployment in dynamic 3D environments.

\begin{algorithm}[!t]
\caption{Algorithm of Uni-Adapter}
\label{algorithm-uni}
\small
\textbf{Input:} Point cloud stream $\{\mathbf{X}_t\}$, encoders $E_P,E_T$ \\
\textbf{Output:} Final logits $\mathbf{s}^{\text{final}}$

\SetKwInput{KwDefinitions}{Definitions}
\KwDefinitions{\\
    \hspace{3mm}$\mathbf{c}_{{k},{n}}$: $n$-th prototype (cluster center) of class $k$,\\
    \hspace{3mm}${b}_{{k},{n}}$: Number of samples accumulated in that prototype,\\
    \hspace{3mm}$K$: Total number of semantic classes (from text prompts),\\
    \hspace{3mm}$N$: Max number of prototypes per class,\\
    \hspace{3mm}$E_{t}$: Text encoder, \ $E_{p}$: Point cloud encoder.\\
    \hspace{3mm}${r}$: Generic prompt , \ ${y_i}$: $i$-th class name \\
    \hspace{3mm}${\mathcal{H}}$: Entropy function \\
}
\BlankLine

{\textbf{Initialize prototypes:}} \\
Set all $\mathbf{c}_{k,n}\leftarrow 0$ and $b_{k,n}\leftarrow 0$, $N_{k}\leftarrow 0$  for 
all classes $k=1\dots K$ and prototype indices $n=1\dots N$. 
\hfill (each class $k$ can have up to $N$ cluster centers)

\BlankLine
\For{each incoming point cloud $\mathbf{X}_t$}{
    {\textbf{Feature Extraction:}} \\
    $\mathbf{f}_t \leftarrow E_P(\mathbf{X}_t)$ 
    \hfill (compute 3D feature for input point cloud)\\
    $\mathbf{w}_i \leftarrow E_T(\{r,y_i\})$, $i=1\dots K$ 
    \hfill (encode text prompts) \\
    $\mathbf{s}^{\text{main}}_i \leftarrow \mathrm{sim}(\mathbf{w}_i, \mathbf{f}_t)$
    \hfill (compute similarity logits) \\
    $k \leftarrow \arg\max_i \mathbf{s}^{\text{main}}_i$ 
    \hfill (assign pseudo class label)

    \BlankLine
    \textcolor{blue!70}{\textbf{Online Prototyping:}} \\
    \eIf{$N_k < N$}{
        \hspace{1.2mm}$\mathbf{c}_{k,N_k+1} \leftarrow \mathbf{f}_t$ 
        \hfill (initialize new prototype for this class) \\
        $b_{k,N_k+1}\leftarrow 1$ 
        \hfill (first sample in this prototype) \\
        \hfill $N_k\leftarrow N_k+1$ 

    }{

        \hspace{1.2mm}$n \leftarrow \arg\max_j \mathrm{sim}(\mathbf{f}_t, \mathbf{c}_{k,j})$ 
        \hfill (find closest prototype) \\
        $H_t \leftarrow \frac{ \mathcal{H}\left( \text{softmax}\left( \mathbf{s}^{\text{main}} \right) \right) }{ \log(K) } $ \\ 
        $H_{k,n}  \leftarrow \frac{ \mathcal{H}\left( \text{softmax}\left( sim(w_i, c_{k, n}) \right) \right) }{ \log(K) }$ 
        \hfill{(compute entropies)}  \\
        \vspace{-1.5mm}\\
        $\alpha_t=e^{-\beta H_t}$, \ $\alpha_{k,n}=e^{-\beta H_{k,n}}$ 
        \hfill (confidence weights) \\
        $\mathbf{c}_{k,n}^{\text{new}}=
        \frac{\alpha_t \mathbf{f}_t + b_{k,n}\alpha_{k,n}\mathbf{c}_{k,n}^{\text{old}}}
             {\alpha_t+b_{k,n}\alpha_{k,n}}$ 
        \hfill (weighted prototype update) \\
        \vspace{-1.5mm}\\
        $b_{k,n}\leftarrow b_{k,n}+1$ 
        \hfill (increment prototype sample count)
    }

    \BlankLine
    \textcolor{green!50!black}{\textbf{Prototype Reassignment:}} \\
    \vspace{-2mm}\\
    Form $\mathbf{U} = [\mathbf{c}_{1,1}; \dots; \mathbf{c}_{K,N_K}] \in \mathbb{R}^{M \times d}$ 
    \hfill (stack prototypes) \\
    \vspace{-2mm}\\
    $\mathbf{A}=\mathbf{U}\mathbf{U}^\top$ 
    \hfill (pairwise similarity between prototypes) \\
    \vspace{-2mm}\\
    $\hat{\mathbf{A}}_{ij} = \mathbf{A}_{ij}\cdot\mathbb{I}(\mathbf{A}_{ij}\ge\gamma)$ 
    \hfill (threshold low similarity) \\
    \vspace{-2mm}\\
    $\mathbf{D} = \mathrm{diag}\left(\sum_j \hat{\mathbf{A}}_{ij}\right) $ 
    \hfill (degree matrix) \\
    $\mathbf{L}_{\text{norm}} = \mathbf{I}-\mathbf{D}^{-1/2}\hat{\mathbf{A}}\mathbf{D}^{-1/2}$ 
    \hfill (normalized Laplacian) \\
    \vspace{-2mm}\\
    $\mathbf{Z}^* = (\mathbf{I} + \lambda_{\text{reg}}\mathbf{L}_{\text{norm}})^{-1}\mathbf{Z}^{(0)}$ 
    \hfill (graph-based smoothing)

    \BlankLine
    \textcolor{blue!70}{\textbf{Cache Logit Calculation:}} \\
    $\boldsymbol{\Lambda} = \text{diag} \left( \left( \frac{1}{\sum_{m=1}^{M} \mathbf{Z}^*_{m,i}} \right)_{i=1}^{K} \right)$ \hfill (normalization matrix)\\\vspace{-2mm}
    $\mathbf{s}^{\text{cache}} =
    \boldsymbol{\Lambda}\, \mathbf{Z}^{*\top}(\mathbf{U}\mathbf{f}_t)$
    \hfill (compute cache logits)
    \vspace{0.2cm}
    \BlankLine
    {\textbf{Entropy-Based Fusion:}} \\
    \vspace{-2mm}\\ 
    $H_t = \mathcal{H}(\mathrm{softmax}(\mathbf{s}^{\text{main}}))$, \quad
    $H_{\text{cache}} = \mathcal{H}(\mathrm{softmax}(\mathbf{s}^{\text{cache}}))$ \\
    \vspace{-1.5mm}\\
    $\mathbf{s}^{\text{final}} =
    \frac{H_{\text{cache}}\mathbf{s}^{\text{main}} + H_t\mathbf{s}^{\text{cache}}}{H_{\text{cache}}+H_t}$
    \hfill (entropy-based aggregation)
}
\end{algorithm}

\section{Details of the 3D Vision Language Foundation Models}

Uni3D \cite{zhou2023uni3d} is a powerful and scalable foundation model designed for 3D representation learning. It incorporates multi-modal learning by aligning 3D features with image-text representations, enabling effective performance in zero-shot settings.
The architecture is transformer-based, utilizing a pre-trained ViT backbone while replacing the traditional patch embedding mechanism with a specialized point tokenizer. Instead of dividing images into square patches, Uni3D tokenizes point clouds into local regions using Farthest Point Sampling (FPS) and K-Nearest Neighbors (KNN). A lightweight PointNet \cite{qi2017pointnet} module then encodes each region.
The model scales across different configurations by adjusting the number of layers, hidden dimensions, and attention heads. Smaller models, such as Uni3D-Tiny and Uni3D-Small, prioritize efficiency, while Uni3D-Large and Uni3D-Giant push the boundaries of large-scale 3D understanding with billions of parameters.

\noindent
In this work, we employ the large-scale Uni3D-Large model as our point cloud encoder. For the image encoder, we utilize EVA-02-CLIP-E/14+ EVA-CLIP. To facilitate multimodal feature extraction, the point cloud encoder is aligned with both the image and text encoders, enabling a unified representation across different modalities. We also conduct experiments on other large-scale 3D VLFMs, such as ULIP-2 and OpenShape. These models similarly train scalable 3D-native encoders aligned with frozen CLIP image/text encoders. For both ULIP-2 \cite{ulip2} and OpenShape \cite{openshape}, we follow the implementation details outlined in Point-Cache \cite{Point-Cache}. All models utilize publicly available pre-trained  weights from their respective GitHub repositories.

\clearpage

\section{Dataset Details}

\noindent \textbf{ModelNet-40C} \cite{modelnet40c} is a benchmark for robust point cloud classification. This dataset involves applying 15 synthetic corruptions to the test set of ModelNet40 \cite{wu15modelnet}, featuring 2,468 samples across 40 classes. The corruptions include density variations, noise, and geometric transformations, each applied across five severity levels. We refer readers to the official repository for further details. In all of our experiments, we used all corruption types with the maximum severity level set to 5.

\vspace{0.1cm}
\noindent \textbf{ShapeNet-C}~\cite{mirza2022mate}, derived from ShapeNetCore-v2~\cite{chang2015shapenet}, contains synthetic shapes across 55 categories and is designed for robust point cloud classification. It applies 15 corruption types—identical to those used in ModelNet40-C—to simulate real-world conditions, challenging models to generalize to imperfect 3D sensor data. Each corruption type is applied to the test samples of the clean ShapeNetCore-v2 dataset, which contains 10,225 shapes. This dataset introduces greater intra-class variation, making it more challenging for domain adaptation tasks.

\vspace{0.1cm}
\noindent \textbf{ScanObjectNN-C} \cite{mirza2022mate} introduces corruptions to the test set of the ScanObjectNN dataset~\cite{uy2019revisiting}. Unlike synthetic datasets, ScanObjectNN is a real-world benchmark comprising 2,309 training samples and 581 test samples of 3D objects across 15 categories. These real-world samples exhibit greater variation in object appearance, background clutter, and occlusions. ScanObjectNN-C further applies additional corruptions and environmental variations—similar to those in ModelNet40-C—to each sample in ScanObjectNN, more effectively simulating severe real-world noise conditions.

\vspace{0.1cm}
\noindent \textbf{Objaverse-LVIS} \cite{Objaverse} is a large-scale benchmark curated to support research in 3D vision and language understanding. It comprises a cleaned subset of data containing over 46,000 high-quality 3D shapes categorized into 1,156 diverse object classes. By covering a broad spectrum of everyday objects, it facilitates evaluation in open-world and long-tail recognition scenarios. This benchmark presents a challenging testbed for measuring the scalability of our Uni-Adapter.

\vspace{0.1cm}
\noindent \textbf{OmniObject3D}~\cite{OmniObject3D} provides a large-scale benchmark designed to advance 3D understanding in open-world settings, featuring diverse scenes and rich multi-view imagery. The dataset comprises 216 object categories with precise 3D annotations, offering a challenging and comprehensive resource that tests generalization to new classes.

\section{Baseline Details} \label{sec:baselines}  
\label{sup:baseline_detail}

In this section, we provide an overview of key TTA methods, highlighting their core mechanisms and essential hyperparameters.

\vspace{0.1cm}
\noindent
\textbf{Test-Time Entropy Minimization (TENT)}: TENT~\cite{wang2020tent} optimizes model adaptation by minimizing prediction entropy via an unsupervised loss function during inference. Like PL, its effectiveness is governed by the learning rate and batch size. However, its applicability to the Uni3D-Large model is constrained, as TENT primarily relies on batch normalization layers, which are scarce in this architecture, limiting its capacity for effective adaptation. The other hyperparameter of this method is the learning rate.

\vspace{0.1cm}
\noindent
\textbf{Source Hypothesis Transfer (SHOT):} SHOT~\cite{shot}  combines entropy minimization, diversity maximization, and self-supervised pseudo-labeling. It also updates batch normalization statistics based on incoming test batches. Key hyperparameters include learning rate, number of adaptation steps, and the weight assigned to the pseudo-labeling loss.

\noindent
\textbf{Sharpness-Aware and Reliable Entropy Minimization (SAR): } Building upon entropy minimization, SAR~\cite{sar} excludes high-entropy predictions to prevent model collapse and incorporates Sharpness-Aware Minimization (SAM) to enhance adaptability to flat minima. Essential hyperparameters are learning rate, adaptation steps, an entropy threshold for filtering uncertain samples, and an epsilon threshold for computing sharpness in SAM.

\vspace{0.1cm}

\noindent
\textbf{Dynamic Unsupervised Adaptation (DUA): }DUA ~\cite{dua} recalibrates batch normalization (BN) statistics during testing. Initially, BN layers remain in their trained state, but as test batches arrive, the model iteratively updates BN statistics using a moving average approach. The number of adaptation steps and the decay factor governing the moving average update rate are crucial hyperparameters.

\vspace{0.1cm}

\noindent
\textbf{Laplacian Adjusted Maximum-Likelihood Estimation (LAME) : } LAME~\cite{lame} modifies model predictions during testing by adjusting prediction probabilities to enforce similarity among related samples in the feature space. Critical hyperparameters include the kernel affinity parameter (defining the kernel density function), the number of nearest neighbors considered, and the maximum number of iterations for optimizing output probabilities.

\vspace{0.1cm}

\noindent
\textbf{Test-Time Prompt Tuning (TPT): } TPT~\cite{tpt} adapts vision-language models like CLIP \cite{clip} at test time by optimizing soft prompts through backpropagation on unlabeled test data. These soft prompts are prepended to class names, and their embeddings are tuned to reduce prediction entropy or maximize consistency across augmentations. Unlike training-free methods, TPT involves iterative gradient-based updates during inference, making it more computationally intensive. Key hyperparameters include learning rate, number of adaptation steps, entropy or consistency loss coefficients, and the number of augmentations used during optimization.

\vspace{0.1cm}

\noindent
\textbf{Marginal Entropy Minimization with One Test Point (MEMO): } MEMO~\cite{memo} is a single-instance TTA approach that generates multiple augmented versions of a test instance and minimizes the entropy of their averaged predictions. This method reduces entropy and enforces consistency among predictions from different augmentations. Hyperparameters are the learning rate, adaptation steps, and the number of augmentations applied to each test instance.
\vspace{0.1cm}

\noindent
\textbf{Test-Time Templates Adjuster (T3A): } T3A~\cite{T3A} is a training-free test-time adaptation method that refines the classifier during inference by adjusting class decision boundaries based on unlabeled test data. Instead of modifying model parameters, T3A updates the linear classifier by constructing pseudo-prototypes for each class using incoming test samples. Predictions are then made based on their proximity to these prototypes, reducing uncertainty and improving adaptation to domain shifts. Key hyperparameters include the number of filters (defining the number of past samples retained for prototype construction).

\vspace{0.1cm}

\noindent
\textbf{Training-free
Dynamic Adapter(TDA):} TDA~\cite{TDA} is a training-free test-time adaptation method designed for vision-language models like CLIP. It maintains two dynamic key-value caches during inference: a positive cache that stores features of confidently predicted samples as pseudo labels, and a negative cache that records uncertain predictions to suppress unlikely classes. These caches are updated on-the-fly as new samples arrive, enabling adaptive refinement of predictions without backpropagation or prompt tuning. Key hyperparameters include the confidence threshold for cache updates, the cache sizes for both positive and negative caches, and the scaling factors used to modulate similarity importance during self-attention.
\vspace{0.1cm}

\noindent
\textbf{CloudFixer:} CloudFixer~\cite{shim2024cloudfixer} is a test-time input adaptation method specifically designed for 3D point cloud recognition under distribution shifts. Unlike conventional test-time adaptation approaches that focus on model updates, CloudFixer transforms test-time inputs to align with the source domain using a pre-trained diffusion model. This method optimizes geometric transformation parameters, including point-wise displacements and rotation matrices, to correct distortions caused by real-world noise, occlusion, and resolution variations. 
Hyperparameters include the step size for geometric transformations and geometric parameters such as rotation magnitude.

\vspace{0.1cm}

\noindent
\textbf{3DD-TTA: }3DD-TTA~\cite{dastmalchi2024test} is a training-free test-time adaptation  method designed for 3D point cloud classification under distribution shifts. Instead of fine-tuning the model parameters, 3DD-TTA adapts corrupted point clouds back to the source domain using a pre-trained latent diffusion model. The method employs a Variational Autoencoder (VAE) to encode the input into a shape latent and latent points, which are perturbed with Gaussian noise. A denoising diffusion process then refines these latent representations, guided by the Selective Chamfer Distance (SCD) to preserve structural fidelity. This enables the test input to better align with the original training distribution, enhancing classification accuracy without modifying the classifier. The key hyperparameter is the number of diffusion timesteps.

\vspace{0.1cm}

\noindent
\textbf{Hierarchical Point-Cloud Cache Adaptation (Point-Cache):} Point-Cache ~\cite{Point-Cache} is a test-time adaptation method designed for 3D vision-language models (3DVLMs) working on point clouds. Inspired by the TDA framework \cite{TDA}, Point-Cache utilizes a hierarchical memory structure that includes both global and local caches. The global cache stores the entire point cloud’s encoded features, while the local cache stores $m$ representative patch centers per cloud, computed via $k$-means clustering on encoder-layer outputs. Key hyperparameters include the number of clusters $m$ for local patches, the size of the global and local caches, and the weighting strategy for ensembling cache-based and model predictions.

\subsection{Baseline Implementation Details}  

\begin{table}[t!]
\centering
\renewcommand{\arraystretch}{1.2}
\setlength{\tabcolsep}{6pt} 
\small 

\resizebox{0.98\columnwidth}{!}{
\begin{tabular}{p{2.1cm}|p{3.3cm}|p{3.8cm}} 
    \Xhline{3\arrayrulewidth}
    \rowcolor{blue!7} \textbf{Method} & \textbf{Hyperparameter} & \textbf{Search Space} \\ 
    \Xhline{2\arrayrulewidth}

    TENT & Learning Rate & $\{10^{-4}, 10^{-3}, 10^{-2}\}$  \\  
    \rowcolor{blue!3} & Adaptation Steps & $\{ 1, 3, 5, 10\}$ \\  \hline

    SHOT & Learning Rate & $\{10^{-4}, 10^{-3}, 10^{-2}\}$  \\  
    \rowcolor{blue!3} & Adaptation Steps & $\{ 1, 3, 5, 10\}$  \\  
    & Pseudo-label Loss Weight & $\{0, 0.1, 0.3, 0.5, 1\}$ \\ \hline

    SAR & Learning Rate & $\{10^{-4}, 10^{-3}, 10^{-2}\}$  \\  
    \rowcolor{blue!3} & Adaptation Steps & $\{ 1, 3, 5, 10\}$  \\  
    & Entropy Threshold & $\{0.2, 0.4, 0.6, 0.8\}$ \\  
    \rowcolor{blue!3} & Epsilon Threshold & $\{0.01, 0.05, 0.1\}$ \\ \hline

    DUA & Adaptation Steps & $\{ 1, 3, 5, 10\}$  \\  
    \rowcolor{blue!3} & Decay Factor & $\{0.9, 0.94, 0.99\}$ \\ \hline

    TPT  & Adaptation Steps & $\{ 1, 2, 3\}$  \\ \rowcolor{blue!3}
    & \# of Augmentations & $\{4, 8, 16, 32\}$ \\ \hline

    MEMO & Learning Rate & $\{10^{-6}, 10^{-5}, 10^{-4}, 10^{-3}\}$  \\  
    \rowcolor{blue!3} & Adaptation Steps & $\{ 1, 2\}$  \\  
    & \# of Augmentations & $\{16, 32, 64\}$ \\ \hline
    LAME& Kernel Affinity & $\{\text{RBF}, \text{KNN}, \text{Linear}\}$ \\  
    \rowcolor{blue!3} & \# of Neighbors & $\{ 1, 3, 5, 10\}$  \\  
    & Max Steps & $\{1, 10, 100\}$ \\ \hline
    T3A & \# of Filters  & $\{1, 20, 50, 100, 200, 300, 400\}$  \\  
    \hline
    CloudFixer*
     &  Step Size & $\{1, 5, 10, 20, 30 , 50\}$ \\  
    \rowcolor{blue!3} & Rotation Parameter & $\{0.01, 0.02, 0.03\}$  
    \\ \hline
    3DD-TTA & Step Size & $\{1, 5, 10, 25, 35\}$ \\   
    \Xhline{3\arrayrulewidth}
\end{tabular}}
\caption{Hyperparameter search space for test-time adaptation baseline methods.}
\label{suplementery:hyperparameters}
\end{table}

\noindent
We note that in training-based TTA methods, only the point cloud encoder is typically updated; however, in the case of TPT, the soft prompts of the text encoder are optimized instead.
For model adaptation techniques such as TENT, SHOT, SAR, and MEMO, we update the statistics of the batch normalization layers. While TENT and SAR exclusively refine the affine parameters of batch normalization layers following established configurations~\cite{wang2020tent, sar}, MEMO optimizes all model parameters. 

\noindent  
These approaches employ an adaptation strategy, where the model continuously updates its parameters without resetting to the pre-trained state after processing each test batch.  
By default, we utilize a batch size of 16 for online TTA baselines. However, our proposed method, Uni-Adapter, along with per-instance TTA frameworks such as MEMO, TDA, and Point-Cache, is designed to function efficiently with a batch size of 1. As T3A does not inherently include a classifier layer, we incorporate CLIP weights as the classifier. For 3DD-TTA and CloudFixer, we leverage their generative capabilities to modify corrupted incoming point cloud samples while maintaining a consistent batch size of 1.
For Point-Cache and TDA, we followed their respective GitHub repositories and adopted their hyperparameters. For the other adapted methods, the hyperparameter search space used in this work is provided in Table~\ref{suplementery:hyperparameters}.

\section{More Experiments}  
\textbf{Effectiveness of Uni-Adapter Against Distribution Shifts on ULIP-2 and OpenShape.} We evaluate the effectiveness of Uni-Adapter on other 3D VLFMs, as presented in Table~\ref{sup:othermethods_accuracy}. Across all corrupted benchmarks, Uni-Adapter consistently outperforms Point-Cache when integrated with both OpenShape and ULIP-2 as source VLFMs. Quantitatively, when paired with ULIP-2, Uni-Adapter achieves improvements of +7.97\% on ModelNet-40C, +3.88\% on ScanObjectNN-C, and +5.38\% on ShapeNet-C over the baseline ULIP-2, and also outperforms Point-Cache on these benchmarks.
 These results highlight the model-agnostic nature of Uni-Adapter and its robust performance across diverse 3D VLFM architectures.

\vspace{0.1cm}
\noindent \textbf{Statistical Significance of Uni-Adapter's Performance Gains.} 
We conduct paired significance tests and report p-values to determine whether Uni-Adapter’s improvements are statistically meaningful.
Table~\ref{sup:pvalues} reports the p-values from statistical significance testing between Uni-Adapter and a set of training-based and training-free baselines across three corrupted point cloud datasets: ModelNet-40C, ScanObjectNN-C, and ShapeNet-C.  Following standard practice, we adopt a significance threshold of $ p < 0.05 $ , under which results are considered statistically significant. Lower p-values indicate stronger evidence that the performance difference is not due to chance.

Across all comparisons, Uni-Adapter shows statistically significant improvements over the baselines. All p-values fall well below the 0.05 threshold, often by several orders of magnitude---for example, Point-Cache (8.04~$\times$~10$^{-7}$ on ModelNet-40C) and SAR (5.86~$\times$~10$^{-6}$ on ScanObjectNN-C). These results confirm that Uni-Adapter consistently yields meaningful improvements over both training-based and training-free test-time adaptation methods under corruption, demonstrating strong generalization and robustness.

\vspace{0.1cm}
\noindent \textbf{Memory Efficiency.}  Table~\ref{tab:memory} presents the memory usage with Uni3D-Large as the source model. We evaluate the memory impact of Point-Cache and Uni-Adapter for zero-shot adaptation across three datasets with varying numbers of classes. As we observe, Objeverse-LVIS exhibits higher memory usage for both methods due to its larger class set, which requires caching more prototypes and results in larger tensor sizes. Overall, both Uni-Adapter and Point-Cache introduce negligible memory overhead to the source model, even across datasets ranging from small- to large-scale class distributions. As shown in Table~\ref{tab:memory}, Uni-Adapter is consistently more memory-efficient than Point-Cache across all datasets. Its memory usage primarily arises from updating the cache and modifying class logits for each input. Notably, Uni-Adapter adds no additional parameters to the source model—only a lightweight cache—enabling fast adaptation with minimal memory cost, making it well-suited for real-world deployment scenarios.

\begin{table}[!t]
   \small 
   \centering
   \resizebox{\linewidth}{!}{ 
   \begin{tabular}{l c c c}
      \toprule
      \textbf{Method} 
        & ModelNet-40C & ScanObjectNN-C & ShapeNet-C  \\
      \midrule
      ULIP-2 (CVPR24) & 45 & 33.05 & 42.22 \\ 
      \ +Point-Cache (CVPR25) & 52.24 & 35.49 & 44.71 \\
      \ \textbf{+Uni-Adapter (Ours)} & \ccol{\textbf{52.97}} & \ccol{\textbf{36.93}} & \ccol{\textbf{47.6}} \\ 
      \midrule
      O-Shape (NIPS23) & 67.43 & 30.1 & 54.35 \\
      \ +Point-Cache (CVPR25) & 70.84 & 37.76 & 56.53 \\
      \ \textbf{+Uni-Adapter (Ours)} & \ccol{\textbf{72.07}} & \ccol{\textbf{41.26}} & \ccol{\textbf{58.2}} \\ 
      \bottomrule
   \end{tabular}
   } 
    \caption{Top-1 accuracy (\%) on corrupted benchmarks using ULIP-2 and OpenShape. Batch size is set to 1 during adaptation. Results are averaged over all corruption types.}
  \label{sup:othermethods_accuracy}
   \vspace{-0.2cm}
\end{table}

\begin{table}[!t]
   \small
   \centering
   \resizebox{\linewidth}{!}{
   \begin{tabular}{cl c c c}
      \toprule
    & \textbf{Method} & \textbf{ModelNet-40C} & \textbf{ScanObjectNN-C} & \textbf{ShapeNet-C} \\
    \midrule
    &Source-Only & \(6.11 \times 10^{-11}\) & \(8.16 \times 10^{-6}\) & \(4.26 \times 10^{-7}\) \\
     \cmidrule(lr){2-5}
    \multirow{5}{*}{\rotatebox{90}{\textbf{Training}}} 
    &TENT  & \(1.49 \times 10^{-8}\) & \(3.75 \times 10^{-7}\) & \(2.69 \times 10^{-5}\) \\
    &SHOT & \(1.76 \times 10^{-8}\) & \(1.60 \times 10^{-6}\) & \(2.54 \times 10^{-5}\) \\
    &SAR  & \(2.25 \times 10^{-8}\) & \(5.86 \times 10^{-6}\) & \(4.78 \times 10^{-6}\) \\
    &DUA  & \(1.39 \times 10^{-8}\) & \(8.84 \times 10^{-6}\) & \(1.85 \times 10^{-5}\) \\
    &MEMO  & \(5.83 \times 10^{-11}\) & \(3.73 \times 10^{-7}\) & \(4.35 \times 10^{-7}\) \\
    &TPT\textsuperscript{*}  & \(1.60 \times 10^{-7}\) & \(2.93 \times 10^{-5}\) & \(3.58 \times 10^{-4}\) \\
     \cmidrule(lr){2-5}
    \multirow{3}{*}{\rotatebox{90}{\textbf{Training-Free}}}     
    &LAME  & \(8.54 \times 10^{-11}\) & \(1.61 \times 10^{-6}\) & \(5.06 \times 10^{-7}\) \\
    &3DD-TTA  & \(1.81 \times 10^{-7}\) & \(2.60 \times 10^{-6}\) & \(3.11 \times 10^{-4}\) \\
    &CloudFixer\textsuperscript{\dag}  & \(1.48 \times 10^{-5}\) & \(6.47 \times 10^{-5}\) & \(5.00 \times 10^{-4}\) \\
    &T3A & \(2.62 \times 10^{-2}\) & \(4.72 \times 10^{-2}\) & \(6.51 \times 10^{-5}\) \\
    &TDA\textsuperscript{*}  & \(2.44 \times 10^{-9}\) & \(7.34 \times 10^{-6}\) & \(3.04 \times 10^{-6}\) \\
    &Point-Cache\textsuperscript{*}  & \(8.04 \times 10^{-7}\) & \(5.45 \times 10^{-4}\) & \(9.66 \times 10^{-8}\) \\
      \bottomrule
   \end{tabular}
   } 
\caption{P-values from pairwise t-tests comparing Uni-Adapter against each baseline method across different corruption benchmarks. }

   \label{sup:pvalues}
   \vspace{-0.2cm}
\end{table}

\begin{table}[!t]
   \renewcommand{\arraystretch}{1.5} 
   \centering
   \resizebox{\linewidth}{!}{ 
   \begin{tabular}{l c c c c}
      \toprule
      \Large \textbf{Method} 
        & \Large ModelNet-40C & \Large Omni3D & \Large O-LVIS  & \Large \#Params (M)\\ 
      \midrule
      \Large Uni3D & \Large \textbf{5,064} & \Large \textbf{5,062} & \Large \textbf{5,062} & \Large \textbf{1,016.5}\\ 
      \ \Large +Point-Cache (CVPR25) & \Large 5,064 & \Large 5,068 & \Large 5,090 & \Large 1,016.5 \\
      \ \Large \textbf{+Uni-Adapter (Ours)} & \Large \ccol{5,064} & \Large \ccol{5,067} & \Large \ccol{5,082} & \Large \ccol{1,016.5} \\ 
      \bottomrule
   \end{tabular}
   } 
    \caption{Comparison of memory usage (in MB) between the Uni-Adapter and Point-Cache. All experiments were conducted with a batch size of 1 on an RTX 4090 GPU.}

   \label{tab:memory}
   \vspace{-0.2cm}
\end{table}

\begin{table}[h!]
\centering
\setlength{\tabcolsep}{3pt} 
\renewcommand{\arraystretch}{1.5} 
\resizebox{1.0\columnwidth}{!}{
\begin{tabular}{c c c c c c c c}
\Xhline{1pt}
\# & \begin{tabular}{c} Online\\ Prototyping \end{tabular}  
& \begin{tabular}{c} Prototype \\ Reassignment \end{tabular}  
& ModelNet40 & SONN & ShapeNet &  O-LVIS &  Omni3D \\ 
\midrule
0 & \large \ding{55} & \large \ding{55} &  78.84 &  59.55 &  79.9 & 46.2 & 32.06 \\
1 & \Large \checkmark & \large \ding{55} & \large 70.13 & \large 61.42 & \large 80.88 & 46.98& 35.37\\
2 & \Large \checkmark & \Large \checkmark & \large \ccol{\textbf{83.96}} & \large \ccol{\textbf{64.03}} &  \ccol{\textbf{81.23}}  &\ccol{\textbf{47.49}} &\ccol{\textbf{35.95}} \\
\Xhline{1pt}
\end{tabular}
}
\caption{Ablation study of different components of Uni-Adapter on clean datasets. The metric used is top-1 classification accuracy. Row \#0 reports the performance of the source model without adaptation. The results are averaged over all corruption types. SONN refers to ScanObjectNN.}

\label{sup: ablation-component-clean}
\end{table}

\begin{figure}[t!]
	\centering
	\captionsetup{skip=5pt} 
	\includegraphics[width=1
    \linewidth]{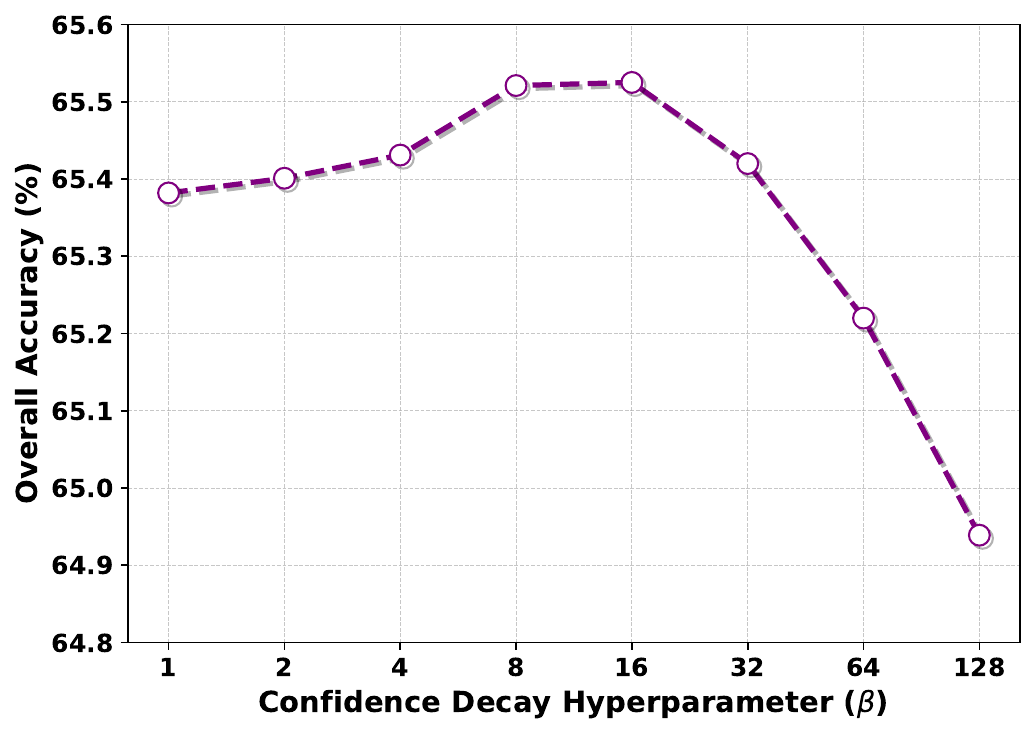}
\caption{Performance with respect to the confidence decay hyperparameter ($\beta$) on the combined ModelNet-40C and ShapeNet-C datasets.}
	\label{sup:beta}
\end{figure}

\section{Ablation Study}  
\label{sup:tab:pvalues}
This section provides a detailed analysis of Uni-Adapter components and their hyperparameter impact.

\vspace{0.1cm}
\noindent \textbf{Ablation of Uni-Adapter Components on Clean Dataset.} Beyond evaluations on corrupted datasets, we observe that both the online prototyping module and prototype reassignment contribute to improved generalization on clean data. As shown in Table~\ref{sup: ablation-component-clean}, the online prototyping mechanism captures intra-class variance and facilitates similarity-based decisions, thereby reducing the modality gap and enhancing generalization to unseen data. Furthermore, in row \#2, our Uni-Adapter mitigates the impact of noisy pseudo-labels during zero-shot inference using the Uni3D source model. Leveraging both components, Uni-Adapter achieves substantial improvements in generalization performance on unseen test samples.

\vspace{0.2cm}

\noindent \textbf{Effect of the Confidence Decay Hyperparameter ($\beta$).} Beta controls the trade-off between confidence and diversity in the online clustering algorithm. When beta approaches zero, the algorithm prioritizes diversity, resulting in a typical clustering approach. Conversely, due to its exponential formulation, increasing beta leads to the selection of only high-confidence samples for prototyping, which significantly influences the final logits during entropy-weighted aggregation. Figure~\ref{sup:beta} illustrates the impact of varying beta values on the combined ShapeNet-C and ModelNet-40C datasets. Our findings suggest that a moderate value of $\beta$, such as 10, strikes an optimal balance between diversity and confidence.

\begin{table}[!t]
\centering
\setlength{\tabcolsep}{3pt} 
\renewcommand{\arraystretch}{1.5} 
\resizebox{1.0\columnwidth}{!}{
\begin{tabular}{c c c c c c}
\Xhline{1pt}
\# & \begin{tabular}{c} Direct \\ Inverse \end{tabular}  
& \begin{tabular}{c} Conjugate \\ Gradient  \end{tabular}  
& ModelNet-40C & ScanObjectNN-C & ShapeNet-C  \\ 
\midrule
0 & \large \ding{55}  & \large \ding{55} & \large 26.61 & \large 26.56 & \large 26.89  \\
1 & \large \checkmark  & \large \ding{55} & \large 29.2 & \large 29.01 & \large 29.53 \\
2 & \Large \ding{55}  & \large \checkmark  & \large \ccol{27.07} & \large \ccol{26.98}  & \large \ccol{27.15} \\

\Xhline{1pt}
\end{tabular}
}
\caption{Comparison of the average inference time (in ms) between the direct inverse method and the conjugate gradient solver for computing Equation~(Eq.~9) in the Uni-Adapter. Results are based on the Uni3D source model. \#0 indicates the average inference time using the Uni-Adapter without graph-based smoothing.}

\label{tab:time-graph}
\end{table}

\vspace{0.2cm}
\noindent \textbf{Efficiency of Conjugate Gradient Solver.} To reduce the computational burden of solving Eq. 9 in the Uni-Adapter proposed method, we adopt the conjugate gradient method~\cite{hestenes1952methods}, which efficiently handles large linear systems. This approach leverages the sparsity and positive semi-definite properties of the Laplacian matrix, enabling iterative solutions without expensive matrix factorizations. By operating only on the non-zero entries, the solver scales effectively with graph size and structure.

Using the conjugate gradient method not only accelerates inference but also maintains high solution quality. As illustrated in Table~\ref{tab:time-graph}, we compare the average inference time between the Direct Inverse method and the Conjugate Gradient method during graph-based smoothing. We observe that the Conjugate Gradient method consistently achieves lower inference latency across all corrupted benchmarks. For example, on the ModelNet40-C dataset, avoiding graph-based smoothing in Uni-Adapter results in an average sample processing time of 26.61 ms, which is attributed to online prototyping and cache logit calculation. When using the direct inverse method for graph-based smoothing implemented in PyTorch, this time increases to 29.20 ms. However, using the conjugate gradient method reduces the time to 27.07 ms, providing a significant efficiency improvement for graph-based smoothing. The method also converges rapidly and introduces a negligible mean absolute error (MAE) in practice—remaining below 0.0005\% compared to using exact inverse computation.
This property makes the method particularly suitable for deployment in real-world applications, where balancing computational efficiency and predictive performance is crucial, especially as the size of the sparse graph increases. In our experiments, we set the maximum number of conjugate gradient iterations to 100.

\section{Visualization}
\begin{figure*}[t]
    \centering
    \captionsetup{skip=5pt} 

    \begin{minipage}[t]{0.48\textwidth}
        \includegraphics[width=\linewidth]{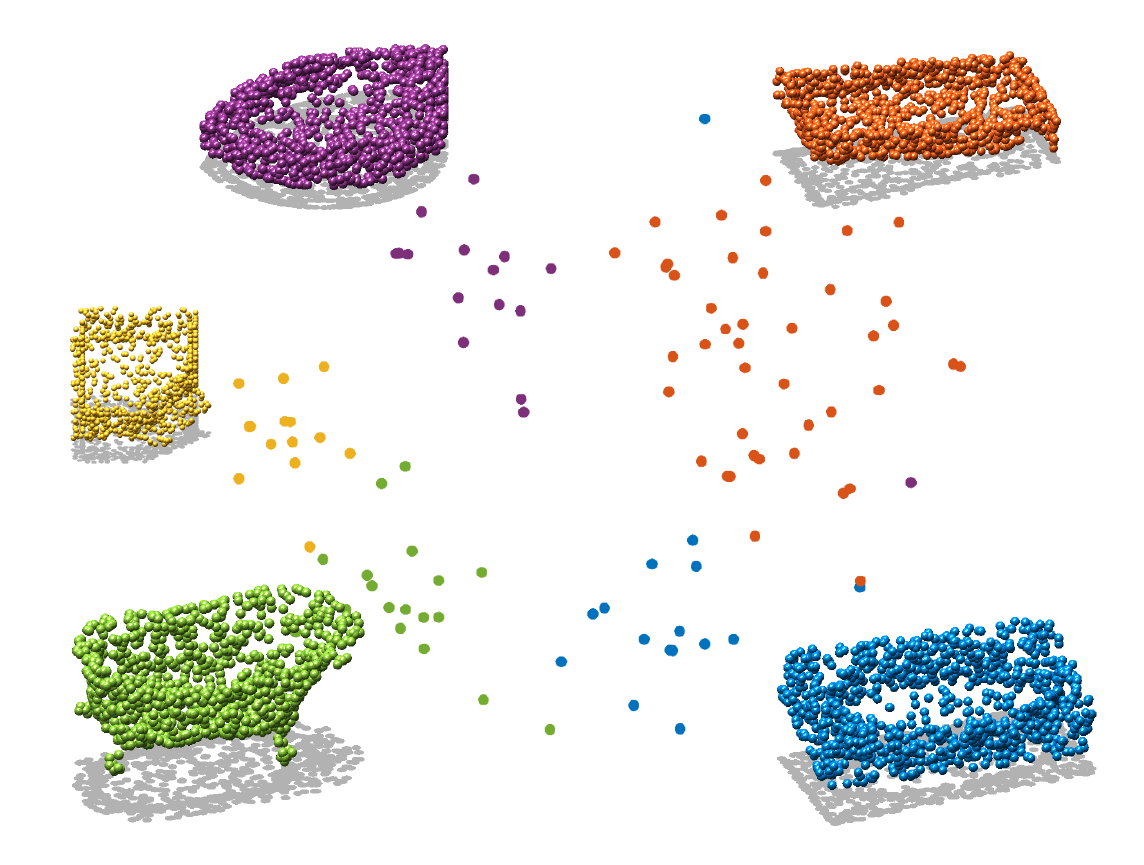}
        \vspace{5pt}
        
        \includegraphics[width=\linewidth]{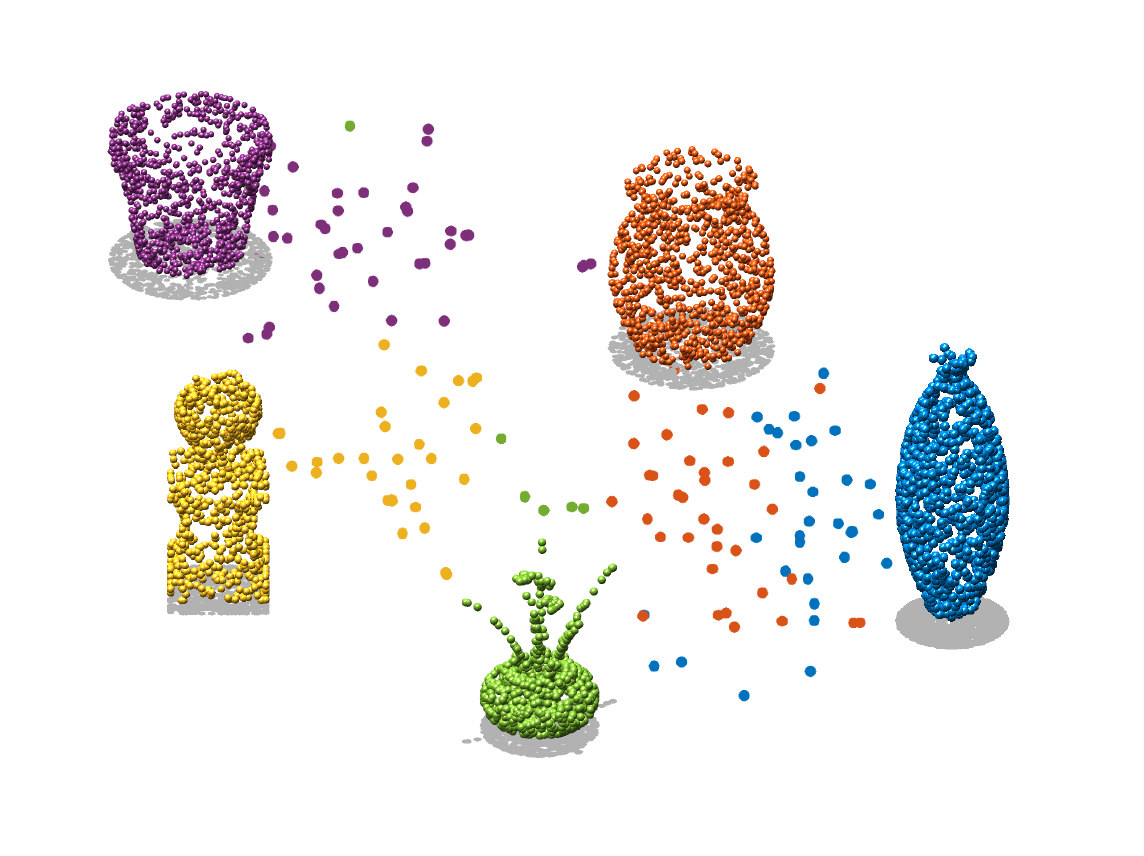}
        \vspace{5pt}

\includegraphics[width=\linewidth]{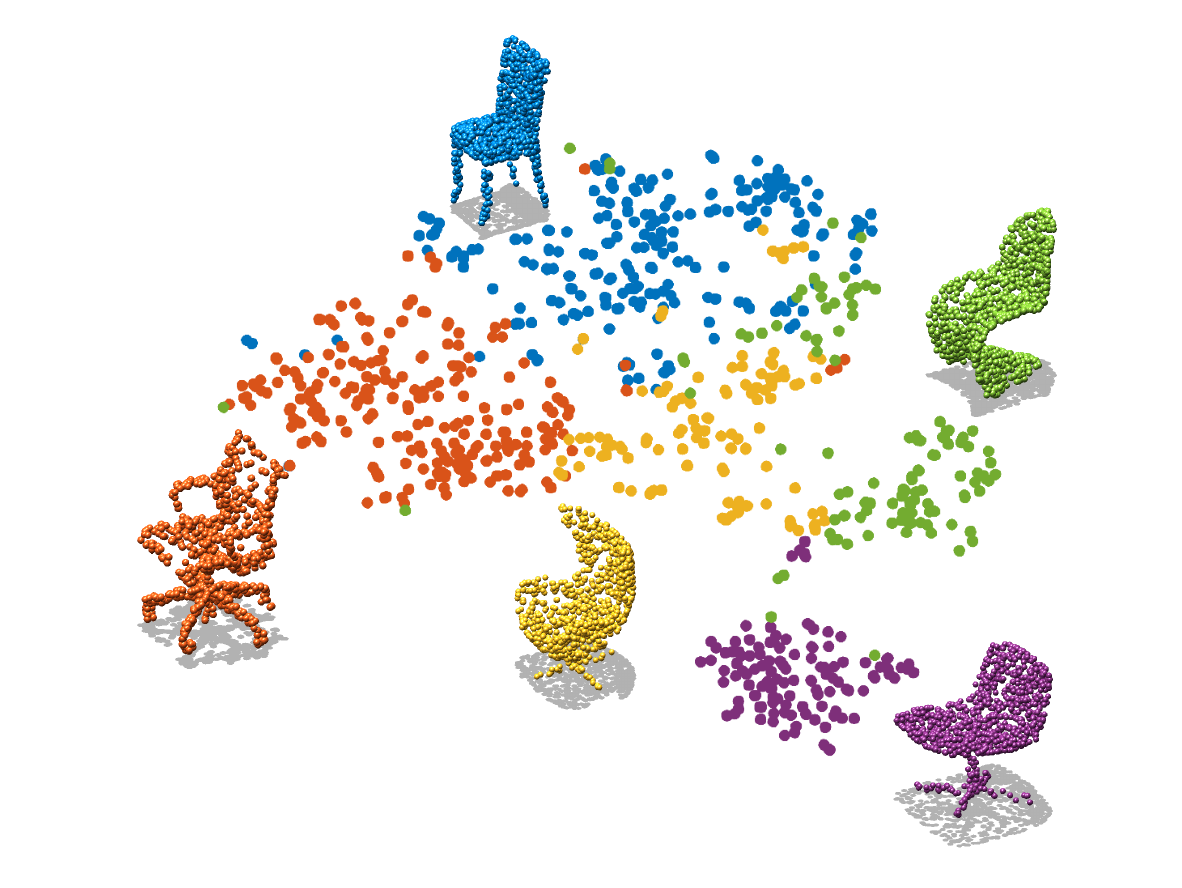}
    \end{minipage}
    \hfill
    \begin{minipage}[t]{0.48\textwidth}
        \includegraphics[width=\linewidth]{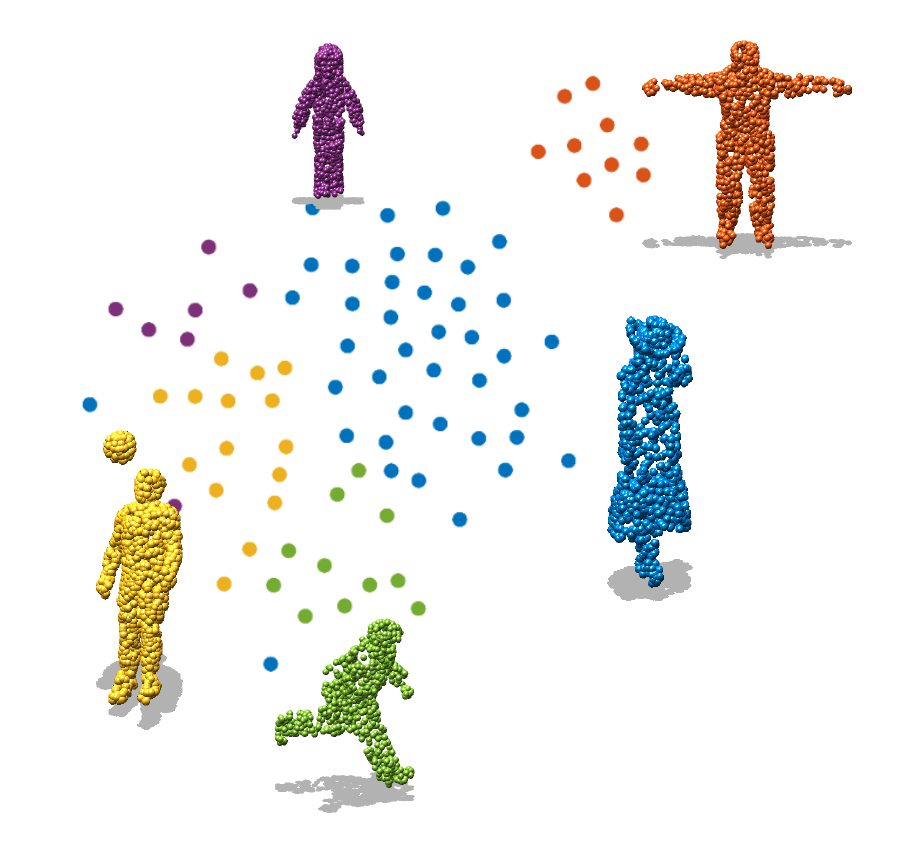}

        \includegraphics[width=\linewidth]{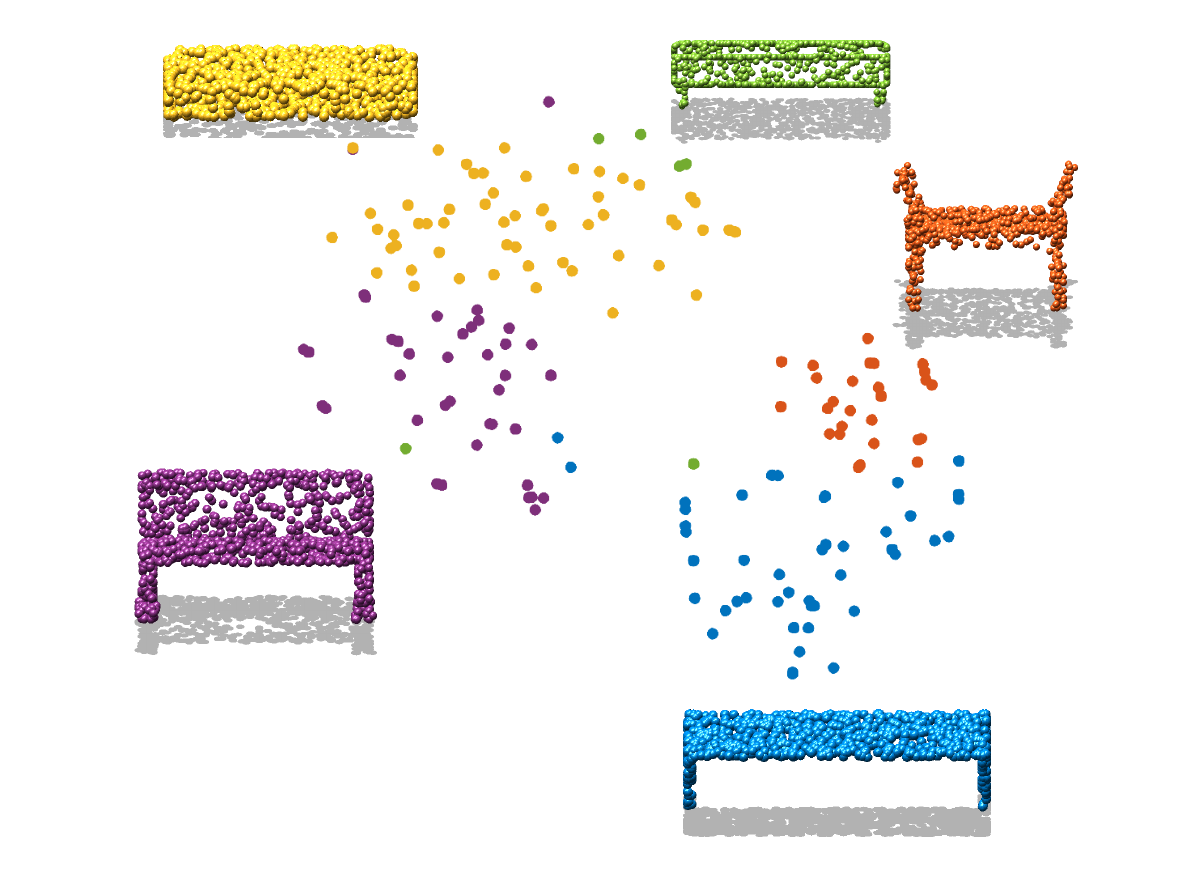}
        \vspace{5pt}
        \includegraphics[width=\linewidth]{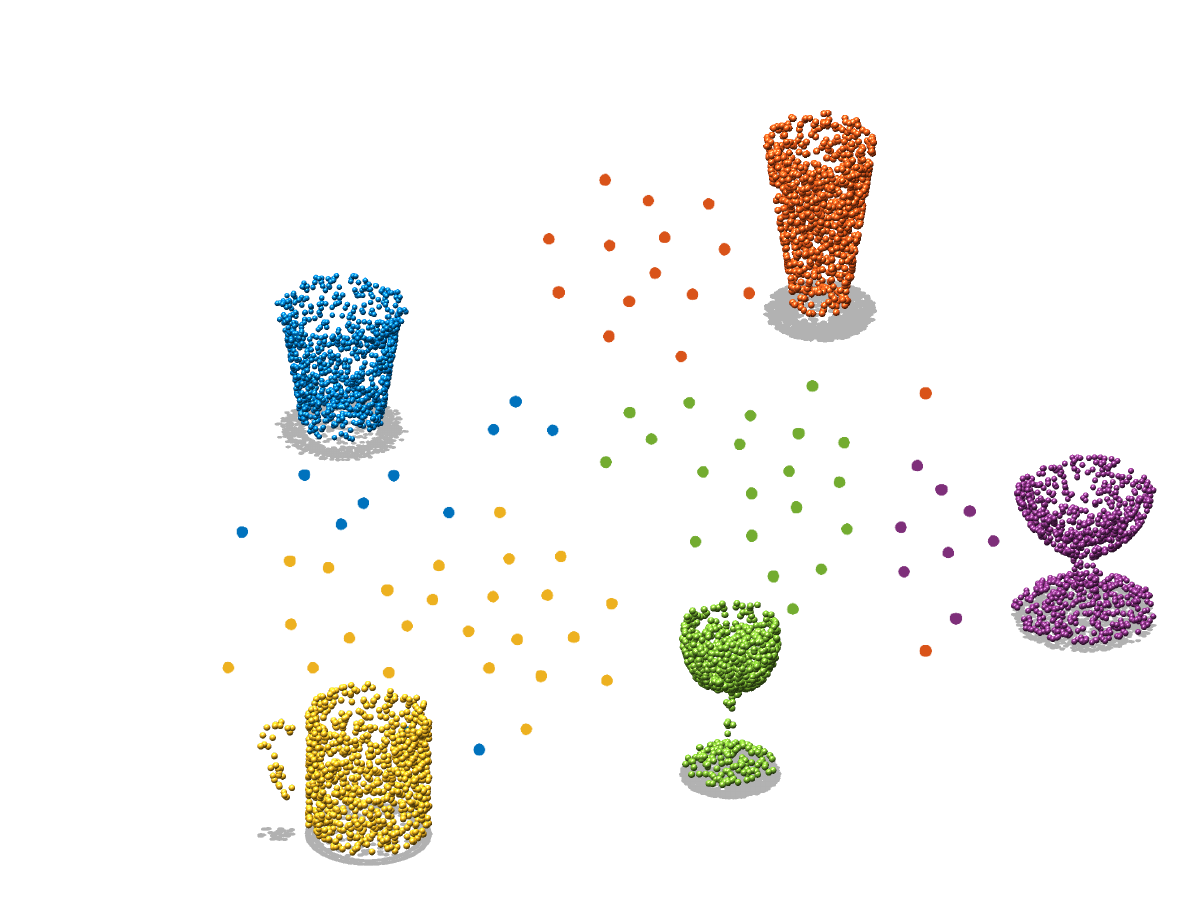}
    \end{minipage}

\caption{Visualization of intra-class variations across different object categories using t-SNE: Bathtub, Bench, Chair, Flowerpot, Person, and Cup.}

    \label{sup:tsne-all_objects}
\end{figure*}

\noindent \textbf{This section visualizes the intra-class variation in 3D point cloud classes.} The visualizations are produced by applying t-SNE to feature embeddings extracted from the Uni-3D source model. These embeddings are obtained using a benchmark created by introducing the same corruptions from ModelNet40-C into the ModelNet40 training set.
Sample instances are drawn from six object categories: Bathtub, Bench, Chair, Flowerpot, Person, and Cup, as visualized in Figure~\ref{sup:tsne-all_objects}.  
As observed, the feature space of these classes exhibits multiple modes and substantial intra-class variance.  This observation motivates our design of a cluster-driven prototyping method to effectively capture all distribution modes.

\clearpage

\clearpage